\documentclass[12pt]{article}
\usepackage{amsmath,amssymb,amstext,dsfont,fancyvrb,float,fontenc,graphicx,subfigure,theorem,hyperref}
\usepackage[margin=1in]{geometry}
\usepackage[utf8]{inputenc}
\usepackage{tikz}

\usepackage[strict]{changepage}
\def\abstractname{Abstract -}   
\def\abstract{\begin{adjustwidth}{1cm}{1cm} \par    \footnotesize \noindent {\bf \abstractname} 
\def\endabstract{ \end{adjustwidth} \smallskip }}

{\theorembodyfont{\itshape}\newtheorem{theorem}{Theorem}[section]}
{\theorembodyfont{\itshape}}
{\theorembodyfont{\itshape}\newtheorem{definition}[theorem]{Definition}}
{\theorembodyfont{\itshape}\newtheorem{lemma}[theorem]{Lemma}}
{\theorembodyfont{\itshape}}
{\theorembodyfont{\rm}}
{\theorembodyfont{\rm}}
{\theorembodyfont{\rm}}
{\theorembodyfont{\rm }}

{\theorembodyfont{\rm }\newtheorem{algorithm}[theorem]{Algorithm}}

\title{\Large\bf A motion planning algorithm in a figure eight track}
\author{\sc C. Jardon, B. Sheppard, V. Zaveri}

\begin{document}

\maketitle

\begin{abstract}
We design a motion planning algorithm to coordinate the movements of two robots along a figure eight track, in such a way that no collisions occur. We use a topological approach to robot motion planning that relates instabilities in motion planning algorithms to topological features of configuration spaces. 
The topological complexity of a configuration space is an invariant that measures the complexity of motion planning algorithms. We show that the topological complexity of our problem is $3$ and construct an explicit algorithm with three continuous instructions. 
\end{abstract}
 

\begin{MSC}
55P99; 05C85; 90C35, 05C90
\end{MSC}

\section{Introduction}
The motion planning problem in robotics relates to the movement of robots from one location to another in their physical space. When there are several robots moving in the same physical space without collisions we need to study the configuration space. 

The configuration space gives a complete specification of the position of every robot in the physical space
and a path in the configuration space is
 interpreted as a particular collision-free movement  from the initial  position of the robots to their final one.

When $X$ is the configuration space, the
motion planning problem consists in constructing an algorithm that takes as input a pair of initial and final
configurations $(x_i, x_f)\in X\times X$, and produces a continuous path $\alpha: I\to X$
from the initial configuration $x_i =\alpha(0)$ to the final one $x_f =\alpha(1)$. Farber initiated a topological approach to this problem in \cite{F1}. Let $PX$ be the space of all paths in $X$. The {\em motion planning algorithm} (MPA) is a map $s : X\times X \to PX $ defined by sending a pair of points to a path between them. This approach favors maps that depend continuously on the input since they will translate into more robust algorithms in real-life problems. However, Farber proved that there exists a continuous MPA in $X$  if and only if $X$ is contractible. Therefore, most MPAs in real life may have essential discontinuities, or instabilities, due to the topology of $X$.

Motivated by this, Farber introduced the {\em topological complexity} of a space. This is a numerical homotopy invariant which measures the minimal number of instabilities of any MPA on a space. Essentially, the topological complexity of a space $X$, $TC(X)$, is the smallest $k$ such that $X\times X$ can be covered by $k$ sets over each of which there is a continuous local section. We will call {\em instructions} each of these local sections. Then, the topological complexity of a space $X$ is the minimal number of instructions in any MPA in $X$.

We present our approach to calculating the topological complexity of the motion planning problem for two robots moving in a figure eight graph, and construct an explicit algorithm with the minimal number of instructions given by the topological complexity.

Our constructive strategy is based on finding a {\em spine} $Z\subset X$ which is a deformation retract of $X$ with lesser dimension and design  the algorithm in $Z$. Then extend the algorithm to the whole configuration space $X$ by using the traces of the homotopy deformation. We also give a translation of the algorithm for the physical space that shows explicitly  how to move each of the two robots from any initial position to any final position.

This paper will be organized as follows. Section \ref{Section2} introduces basic terminology used throughout the paper along with the basic definitions of MPA and configuration spaces. Section \ref{Section3} explains the construction of the configuration space and its flat representation. In section \ref{Section4} we discuss  the topological complexity and its basic properties. Section \ref{Section5} gives a detailed description of the homotopy deformation into the spine. We also show in this section that the spine is homotopy equivalent to a wedge of seven circles, which allows us to calculate the topological complexity of the configuration space. In section \ref{Section6} we present our main algorithm in the spine, and extend it to the whole configuration space.  We translate the previous algorithm in the configuration space to the physical space in section \ref{Section7}.

\section{Definitions and terminology}\label{Section2}

\subsection{Motion planning problem}
One of the goals of robotics is to create autonomous robots. A set of robots is a mechanical system that accepts descriptions of tasks and executes them without further human intervention. The input description specifies what should be done and the robots decide how to do it and perform the task. The {\em motion planning problem} consists of designing a general algorithm for the robots  to move from an initial position to a final one. We will use a topological approach to the robot motion planning problem initiated by Farber in \cite{F1}.

\subsection{Basic notions}
Since continuity is a key feature of our algorithm, we will start by recalling its definition for topological spaces. Let $X$ and $Y$ be topological spaces.

We say that $f : X \to Y$ is {\em continuous at $a \in X$} if whenever $V$ is an open neighborhood of $f(a)$ in $Y$, there is an open neighborhood $U$ of $a$ in $X$ such that $f(U)\subset V$.    If $f$ is  continuous at every point $a\in X$ then we say $f$ is {\em continuous} on $X$.
    
 A {\em path} from a point $A$ to a point $B$ in a topological space $X$ is a continuous function $\alpha$ from the unit interval $I = \left[0,1\right]$ to $X$ with $\alpha(0)=A$ and $\alpha(1)=B$. The {\em path space}, $PX$,  is the space of all paths $\alpha$ in $X$, i.e. $PX = \left\{\alpha \ \lvert \ \alpha : \left[0,1\right] \to X\right\}$. A space $X$ is path-connected if for all points $x,y \in X$ there exists a path from $x$ to $y$.
	
The {\em product space} of $X$ and $Y$ will be the Cartesian product $ X \times Y =  \{ (x,y ) \ \lvert \ x\in X, y\in Y \}$ with the product topology.

 The map $ e: PX \to X\times X$ which takes a path $\alpha$ as its input and returns the pair $(\alpha(0),\alpha(1))$ is called the {\em evaluation map}. Here $\alpha(0)$ is the initial point of the path and $\alpha(1)$ is the final one.

 If $X$ is a path-connected space, let the function $s: X \times X \to PX$ be a section of the evaluation such that $e \circ s = id_{X \times X}$.  This section takes as input two points and produces a path $\alpha$ between them as output: $$ s : (A,B) \mapsto \alpha \text{ and } \alpha(0) = A, \alpha(1)=B$$ where $\alpha(0) = A$ is the initial point and $\alpha(1) = B$ is the final point of the path.

Let $f:X\to Y$ be a bijection. If both the function $f$ and the inverse function $f^{-1} : Y \to X$ are continuous, then $f$ is called a homeomorphism. If such a function exists, we say $X$ is {\em homeomorphic} to $Y$ using the notation $X\approx Y.$

 Let $h$ and $h'$ be continuous maps from $X$ to $Y.$ We say that $h$ is {\em homotopic} to $h'$, denoted by $h \simeq h'$,  if there exists a continuous map $$ F : X \times I \to Y$$ such that $$F(x,0)=h(x) \text{ and } F(x,1) = h'(x)$$ for each $x\in X$.

A map $f:X\to Y$ is a {\em homotopy equivalence} if there exists a map $g: Y\to X$ where $f\circ g \simeq id_Y$ and $g\circ f \simeq id_X.$ If such maps exist, we say $X$ and $Y$ are {\em homotopy equivalent} and write $X\simeq Y$. Note that homeomorphic spaces are always homotopy equivalent.

A topological space $X$ is  {\em contractible} if it is homotopy equivalent to a point.

\subsection{Configuration space}
The variety of all the possible states that any mechanical system can take is called the {\em configuration space}. Each point in the configuration space represents a state of the system. 
The space $\Gamma$ where the robots are able to move will be called the {\em physical space}. The configuration space of $n$ robots moving in a space $\Gamma$ without collisions, denoted $C^n(\Gamma)$, is defined as:
$$C^n(\Gamma)=( \Gamma \times \Gamma \times \cdots \times \Gamma ) - D.$$ Here $D$ represents  the pairwise diagonal: $$D = \left\{(x_1, x_2, \ldots, x_n) \in \Gamma^n \ \lvert \ x_i = x_j \text{ for some } i \neq j \right\}$$
given by the states in which at least two robots occupy the same place. 

\section{Configuration space of two robots on a figure eight track}\label{Section3}
\begin{figure}[H]
    \centering
    \includegraphics[height=1in]{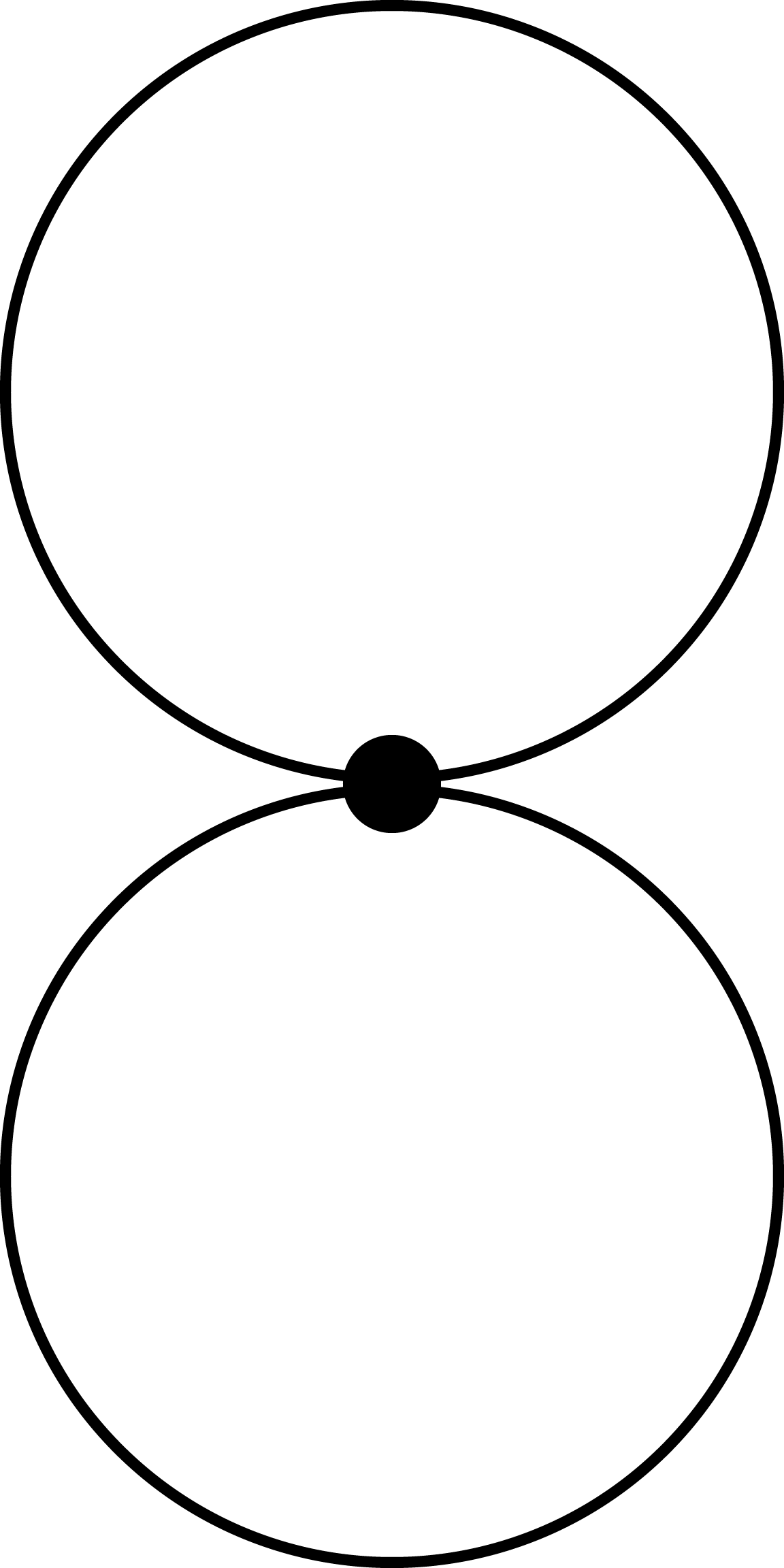}
    \caption{Physical space $\Gamma$.}
    \label{gamma}
\end{figure}

Let $\Gamma$ denote the {\em figure eight} space, which is a wedge sum of two circles $S^1\vee S^1$.  We will consider the motion planning problem of two robots moving in the track $\Gamma=S^1\vee S^1$. See figure \ref{gamma}.

In this case, the configuration space is
$C^2(\Gamma)=( \Gamma \times \Gamma ) - D.$
\noindent We will visualize this space $X=C^2(\Gamma)$ first when embedded in $\mathbb R^3$ and then we will explain its flat representation that will be useful for calculations. 

 \subsection{Visualization of the configuration space $X$}
\begin{figure}[H]
    \centering
    \includegraphics[width=.5in]{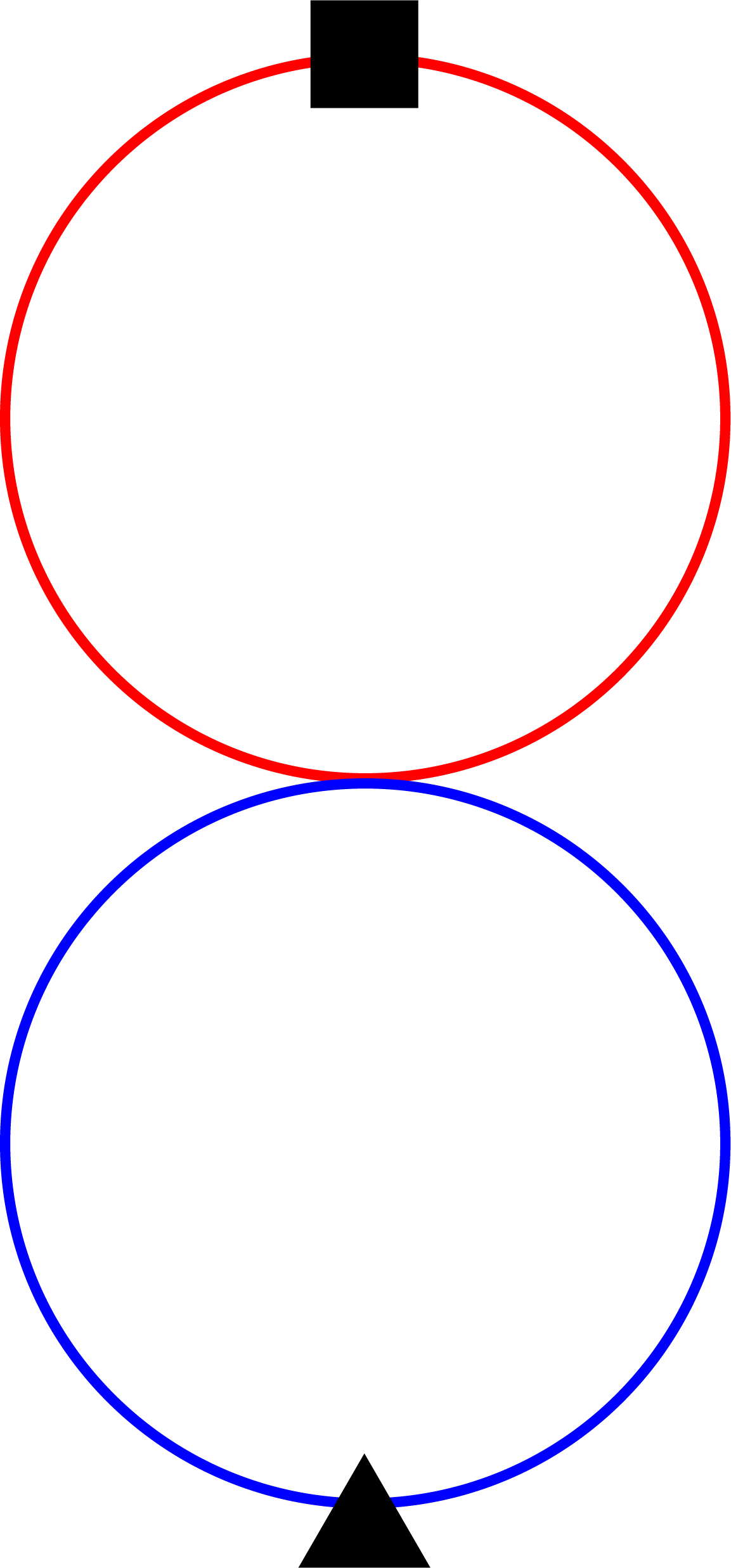}
    \caption{First and second robot in $\Gamma$}
    \label{gammapolepositions} 
\end{figure}

 The robots in $\Gamma$ are represented by triangles and squares. The triangle robot will be also referred to as the {\em first} robot and the square robot as the {\em second} one. We shade each circle with a different color to distinguish the two separate circles in the configuration space. Each concrete position in the physical space, see for instance figure \ref{gammapolepositions}, will have a corresponding state in the configuration space.
            
\begin{figure}[H]
    \centering
    \includegraphics[width=4in,scale=1]{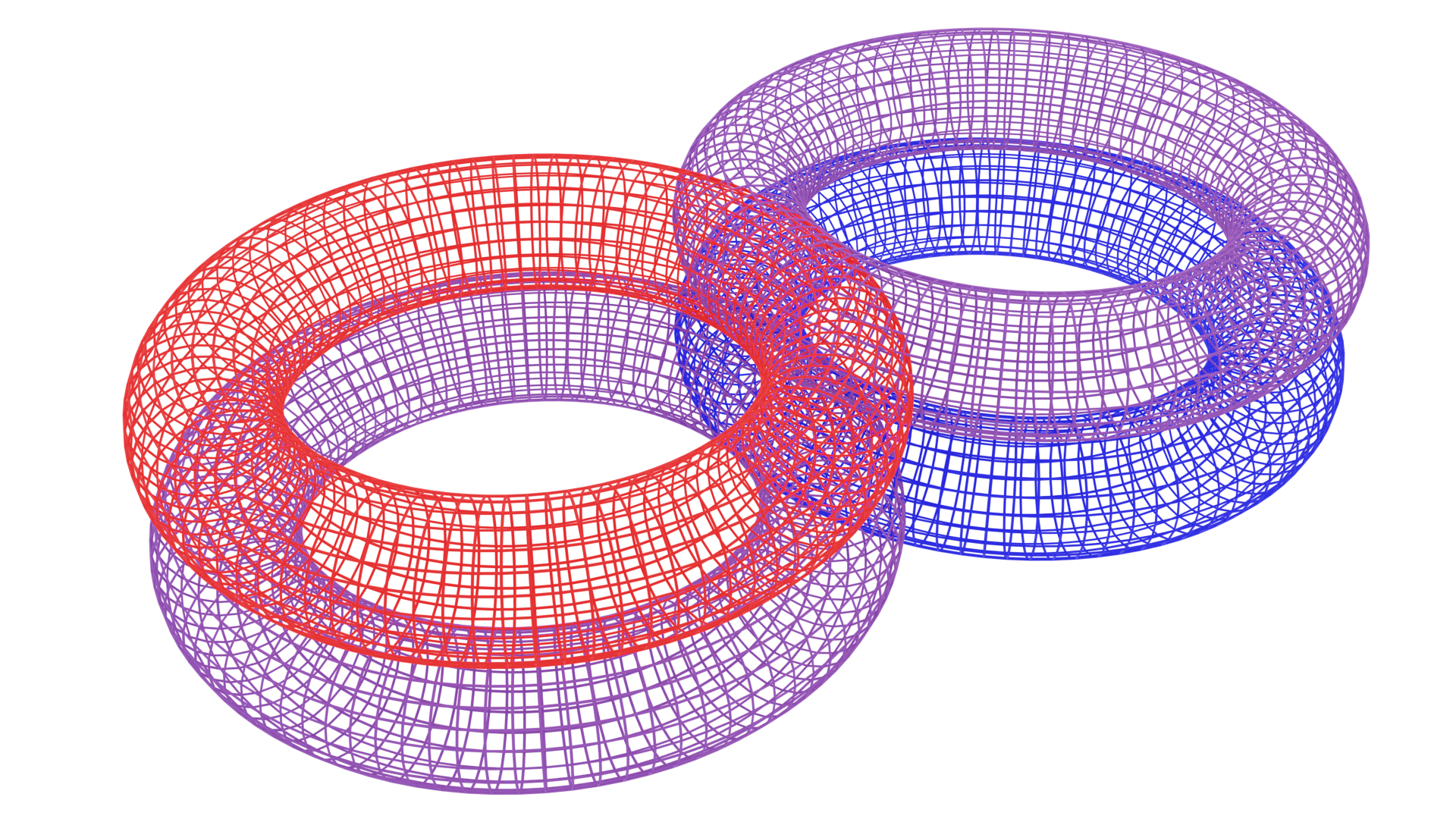}
    \caption{Cartesian product $(S^1\vee S^1)\times (S^1\vee S^1)$.}
    \label{4tori3Dwire} 
\end{figure}

The Cartesian product $\Gamma\times \Gamma$ represents all positions of the two robots in $\Gamma$. The product of a figure eight by itself is given by four connected tori as in figure \ref{4tori3Dwire}.
For robots in the same circle, their states are represented by the red and blue tori whereas the purple tori represent states where the robots are in different circles.

To obtain the configuration space, we need to remove the points in this Cartesian product that correspond to positions in which both robots are at the same place. We observe that the two tori representing positions of robots in the same circle have a diagonal removed from them, whereas the other two tori have just one point removed.

\begin{figure}[H]
    \centering
    \includegraphics[width=4in,scale=1]{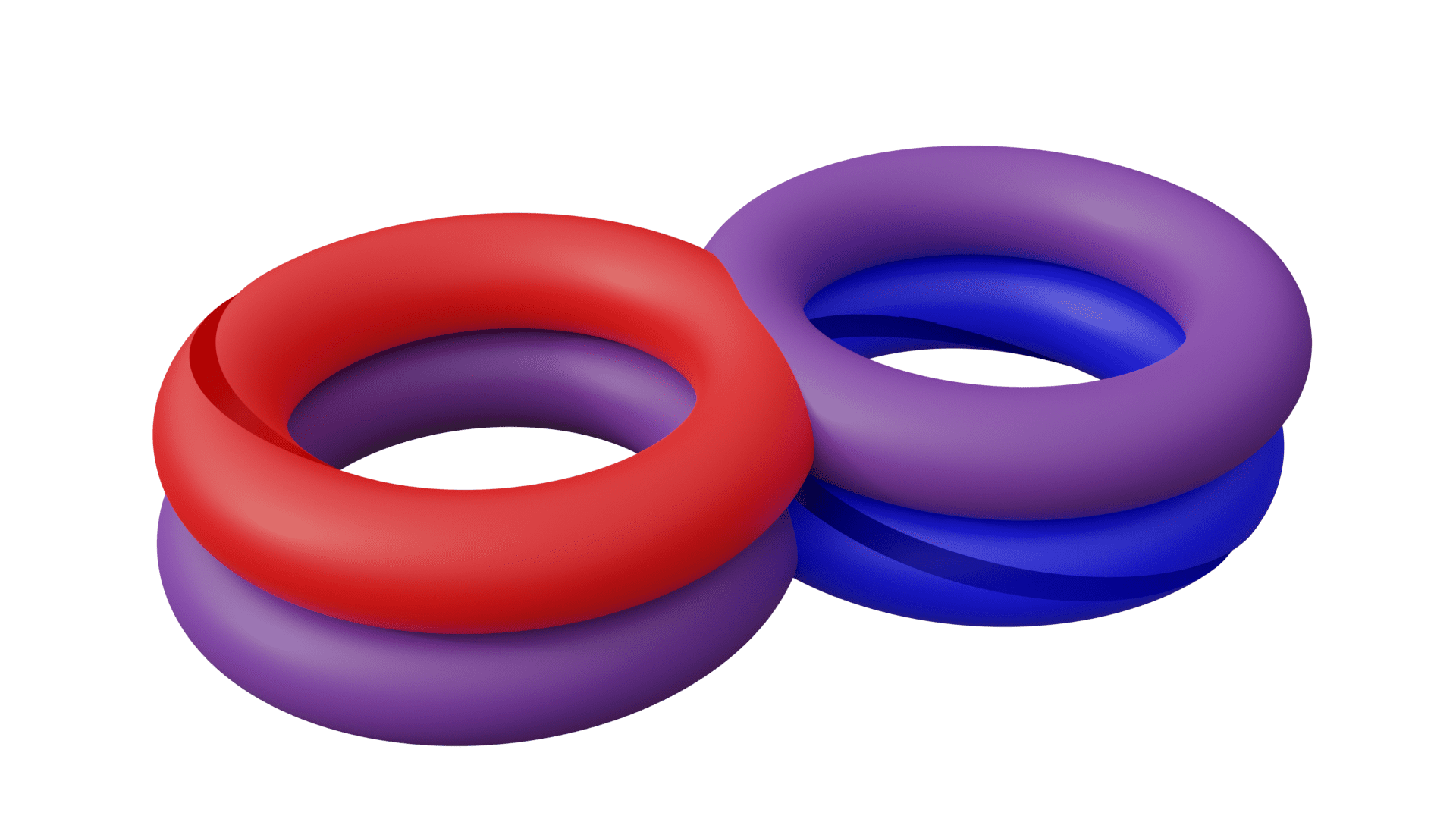}
    \caption{$X=(\Gamma \times \Gamma) - D$.}
    \label{4toriminusd}
\end{figure}

The diagonal $D=\{(x,y)\in (S^1\vee S^1)\times (S^1\vee S^1) |\; x=y\}$ determines a curve on the red and blue tori. The configuration space $X$ is the Cartesian product $\Gamma\times \Gamma$ with this curve removed. See figure \ref{4toriminusd}.

\begin{figure}[H]
    \centering
    \includegraphics[width=2in,scale=0.5]{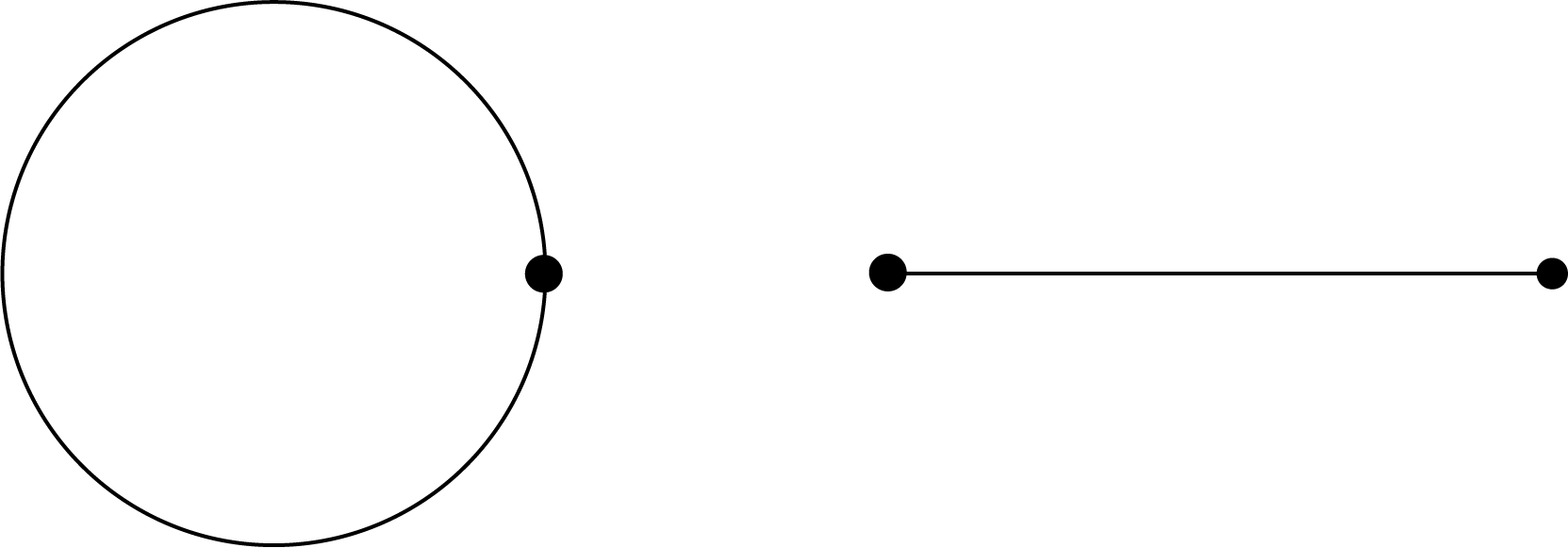}
    \caption{Flat representation of the circle.}
    \label{circleinterval}
\end{figure}

 We will represent now the configuration space in the two-dimensional space by first considering the circle $S^1$ as a quotient of the unit interval where the points $0$ and $1$ are identified, i.e. $S^1\approx I/\{0\sim 1\}$. See figure \ref{circleinterval}.

\begin{figure}[H]
    \centering
    \includegraphics[width=2in,scale=0.5]{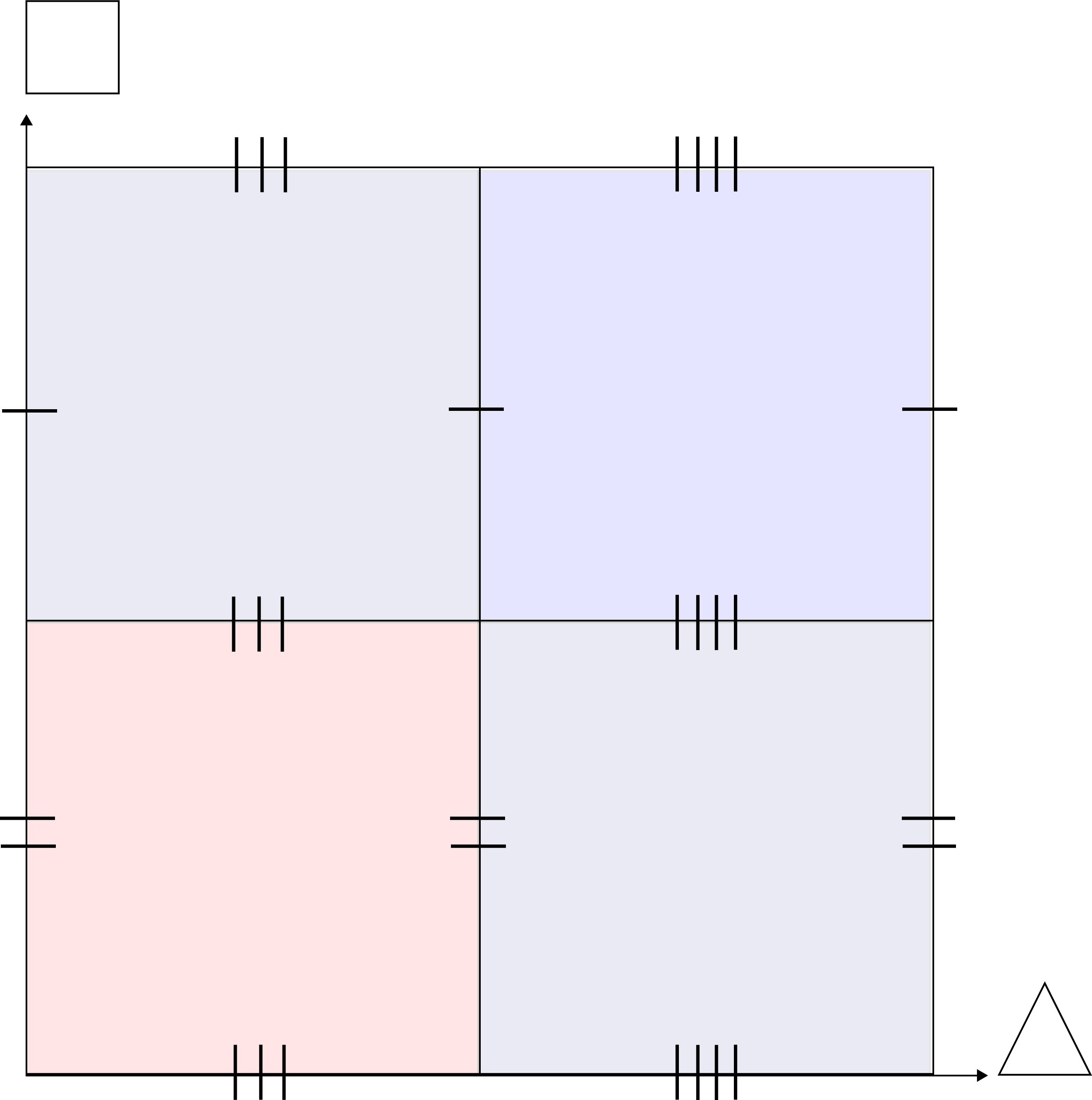}
    \caption{The product $\Gamma\times \Gamma$}
    \label{spacexdashes} 
\end{figure}

The flat representation of the Cartesian product $\Gamma\times \Gamma$  is given by four connected squares with their sides identified as in figure \ref{spacexdashes}.

\begin{figure}[H]
    \centering
    \includegraphics[width=2in,scale=1]{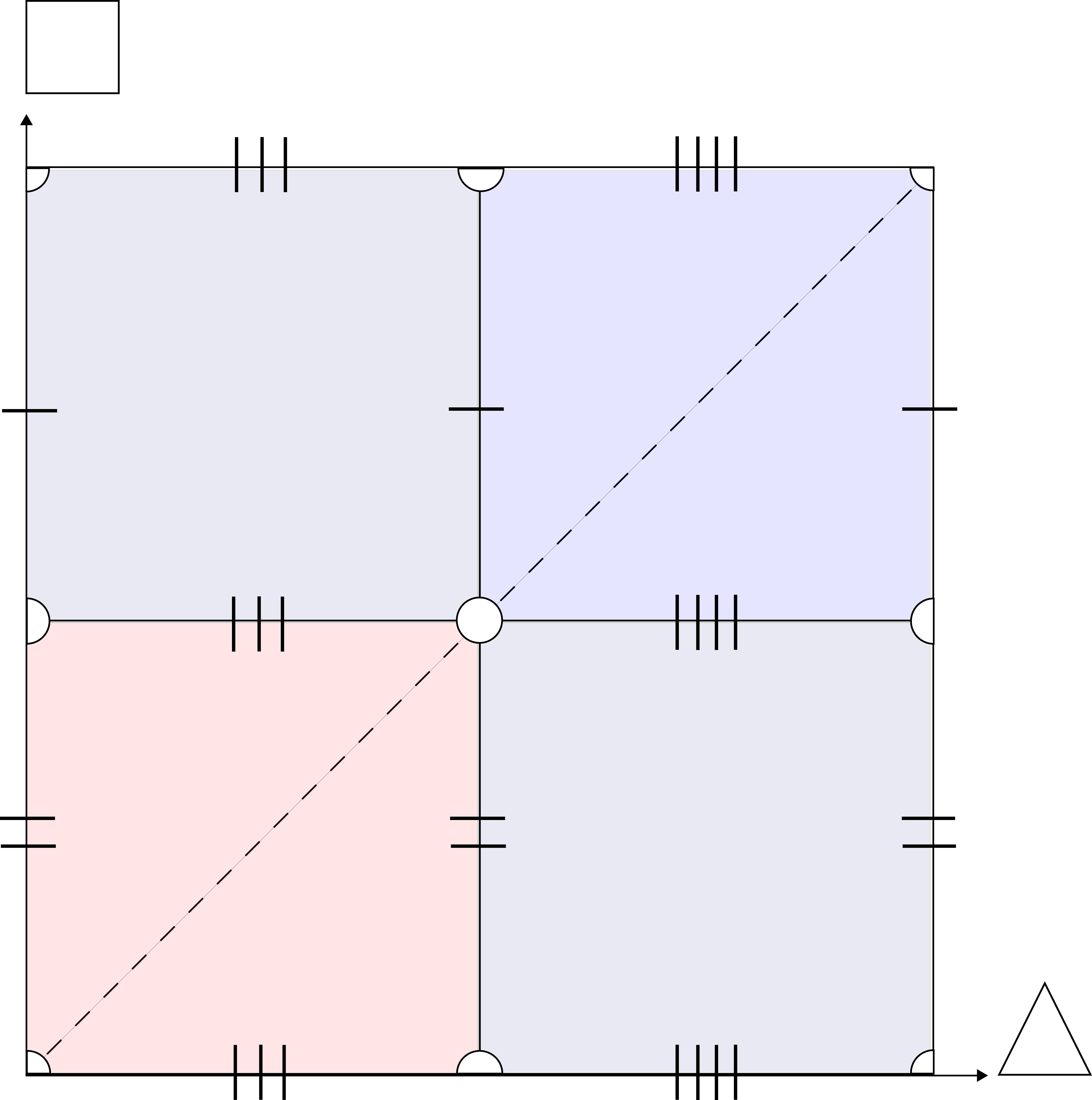}
    \caption{Flat representation of $X=(\Gamma \times \Gamma) - D$}
    \label{xspacediag}
\end{figure}

The diagonal $D=\{(x,y)\in (S^1\vee S^1)\times (S^1\vee S^1) |\; x=y\}$ is given by the diagonals of the red and blue squares in the flat representation.
The configuration space is the Cartesian product $\Gamma\times \Gamma$ with the diagonal $D$ removed. See figure \ref{xspacediag}.

\begin{figure}[H]
    \centering
    \includegraphics[width=5in]{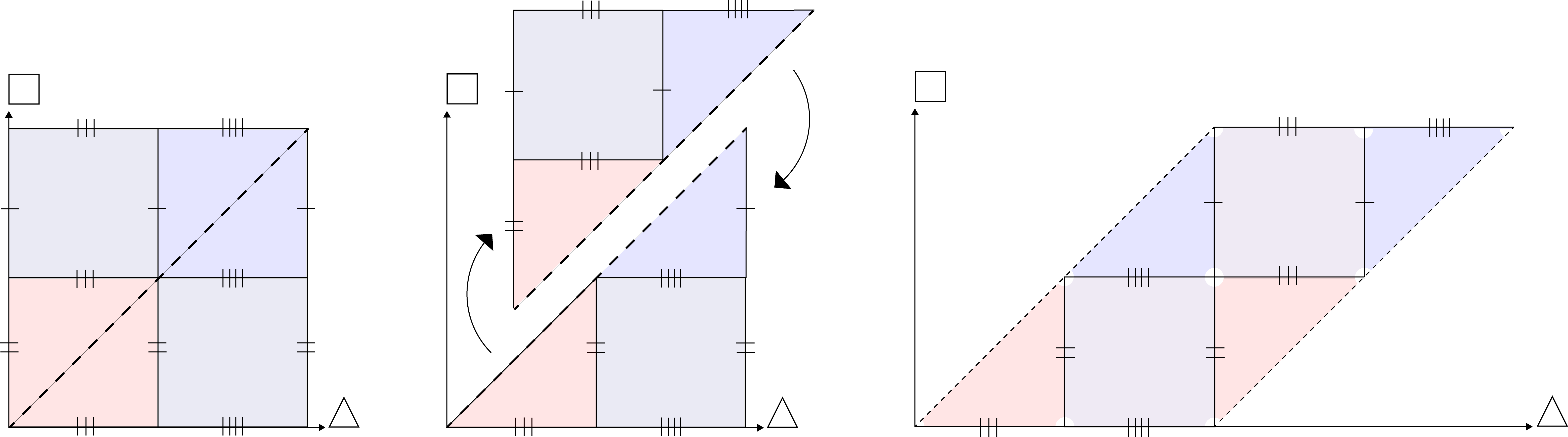}
    \caption{Transition from square to parallelogram representation.}
    \label{transitionparallelogram}
\end{figure}

\begin{figure}[H]
    \centering
    \includegraphics[width=3in]{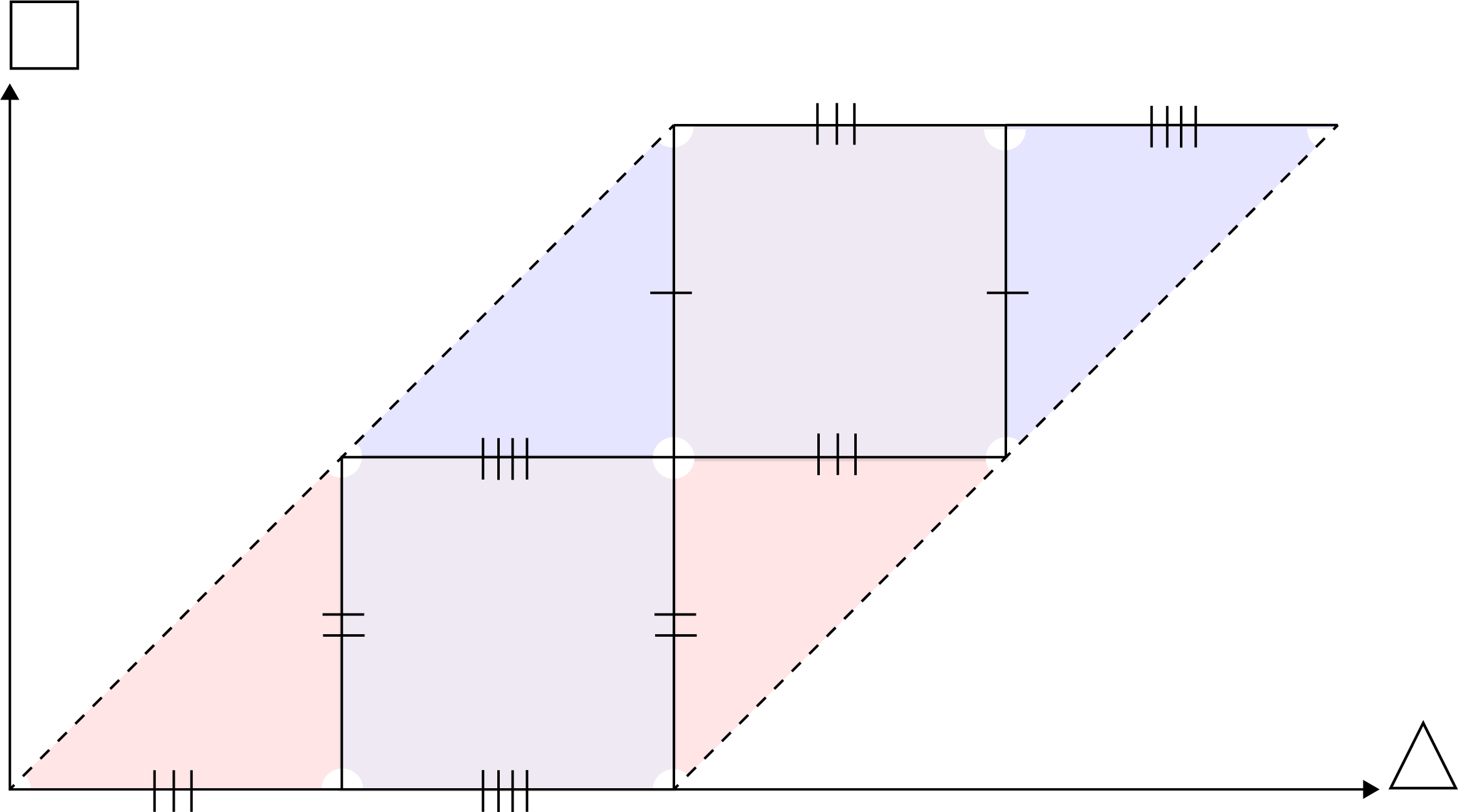}
    \caption{Flat representation of $X=(\Gamma \times \Gamma) - D$ as a parallelogram.}
    \label{fig:paralellogram}
\end{figure}

 Once that the diagonal is removed, we can visualize the flat representation of the configuration space as the following parallelogram with sides identified as shown in figures \ref{transitionparallelogram} and \ref{fig:paralellogram}.

\section{Topological Complexity}\label{Section4}
In order to program robots to move autonomously from any initial state to any desired state we will need a motion planning algorithm that takes as input the pair $(A, B)$ where $A$ is the initial state and $B$ is the final state and 
outputs a continuous motion of the system starting at $A$ and ending at $B$.

           A {\em Motion Planning Algorithm}  (MPA) is a section $s: X \times X \rightarrow PX$ of the evaluation map. It is a function that takes as input a pair of initial and final states, and outputs a path between them. 
           
          A motion planning algorithm $s: X \times X \rightarrow PX$ is a map that is not necessarily continuous. 
          When it is continuous, if the inputs are slightly modified then the path between the inputs is only slightly modified too.  The discontinuities of the MPA  emerge as instabilities of the robot motion.
We want to minimize such instabilities to produce instructions which are robust to noise: if there are small errors in the initial and final positions measurement, we want the paths associated to the exact positions and the one given by the measurement to be nearby.          
          
          Farber proved that the only case where a continuous motion planning algorithm exists is when the space is contractible.

            \begin{theorem}\cite{F1}\label{MPA contractible}
            A continuous motion planning $s$: $X \times X \rightarrow PX$ exists if and only if the configuration space $X$ is contractible.
            \end{theorem}

            For spaces that are not contractible, all MPAs will be discontinuous. In real-life cases, we would like to minimize these discontinuities to produce optimally stable MPAs.
            
            Farber invented a number known as {\em Topological Complexity}, $TC(X)$, that roughly can be described as the minimal number of continuous instructions which are needed to describe any motion planning algorithm in $X$.            
            
                                    This number measures how complex it is for the robots to navigate the space. For example, if the topological complexity of a space is two, then robots need a minimal number of two continuous instructions to move. The higher the $TC$ is, the harder it is for the robots to navigate the space.

            \begin{definition}\cite{F1}\cite{Jose}\label{TC Definition}
            Let $X$ be a path-connected topological space. The {\em topological complexity} of  $X$, $TC(X)$, is the smallest number $k$, such that  $X\times X$ can be covered by $k$ sets $U_1, U_2, U_3$, $\ldots$, $U_k$ where on each of them there is a continuous local section $s_i : U_i \to PX$ for each $i = 1, 2, \ldots, k$.                    
            
            \end{definition}
            Each of these local sections will be our continuous instructions in the motion planning algorithm. 
            For instance, if the space $X$ is contractible we need only one section and $TC(X)=1$.
   In the case of a circle, we need two sets to cover $S^1 \times S^1$. The first set is $U_1 = \{(x,y) |\; x \text{ is antipodal to } y\}$ and the second set is $U_2 = \{(x,y) |\; x\text{ is not antipodal to } y\}$. The first instruction over the set $U_1$ is {\em ``move counterclockwise''} while the second instruction over $U_2$ is {\em ``move following the shortest path''}. Both of these instructions are continuous and $TC(S^1)=2$. Note that none of these instructions are continuous over $S^1 \times S^1$.
              
       Farber also proved that this number is a homotopy invariant. If a space is homotopy equivalent to another, then their topological complexities are the same on their domains.             
            
        \begin{theorem}\cite{F1}\label{invariant}
        If $X$ is homotopy equivalent to $Y$, then $TC(X)= TC(Y)$.
        \end{theorem}

\section{Homotopy type of $X$}\label{Section5}
Our aim will be to find a simpler space that is homotopy equivalent to the space $X$ and then construct our algorithm in the simpler space. Finding an algorithm and controlling the discontinuities directly in $X$ is not trivial. Theorem \ref{invariant} is crucial to reduce the complexity of the task by using the homotopy equivalence not only to calculate the topological complexity of a simpler space, but also to construct the actual algorithm in the simpler space and extend it to $X$ afterward using the homotopy traces.

Ghrist proved that the configuration space of robots moving in a graph is homotopy equivalent to a space of lesser dimension.

\begin{theorem}\cite{F2}\label{dimension}
        Given a graph $\Gamma$, the configuration space of $N$ distinct points on $\Gamma$ can be deformation retracted to a spine whose dimension is bounded above by the number of vertices of  $\Gamma$ of valency greater than two.
\end{theorem}
Our figure-eight graph has only {\em one} vertex with valency greater than two and therefore the configuration space retracts to a spine of dimension $1$. We know then that it is possible to shrink the space $X$ to a one-dimensional space. We will show next the explicit construction of this spine for our graph.

\subsection{Construction of the Spine}\label{traces}

The {\em spine} $Z$ of the configuration space is a one-dimensional space that carries all of the topology of the configuration space through reducing the full space by deformation retraction to a simpler space. The {\em center}  of the physical space $\Gamma$ is the point at which the circles intersect and the {\em poles} are the antipodal points to the center in each circle.

\begin{figure}[H]
    \centering
    \includegraphics[width=1in]{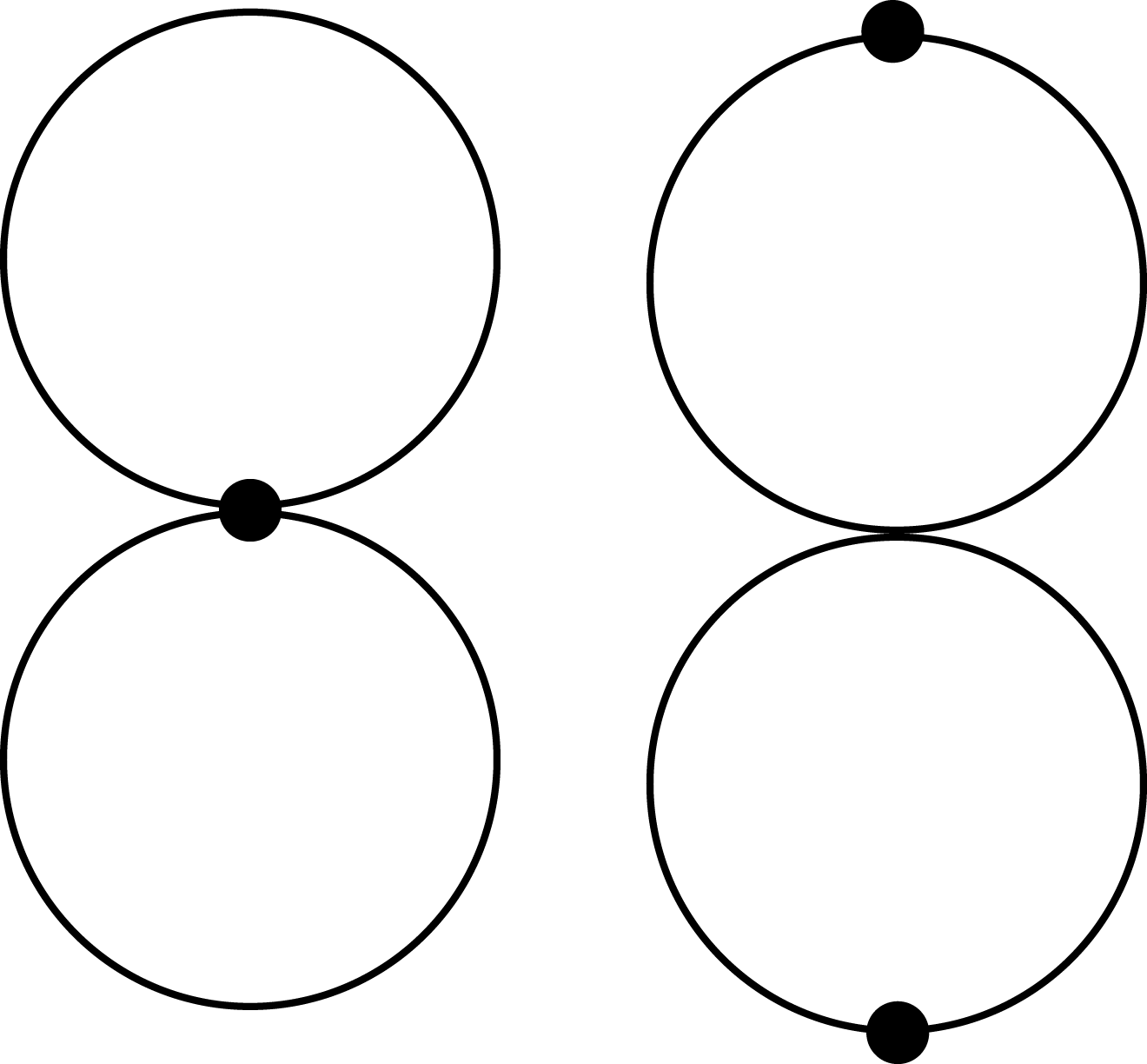}
    \caption{The center and the poles, respectively.}
    \label{gammacenter}
\end{figure}
 
The spine  $Z$ is a set of segments in the flat representation with their extremes identified that represents the positions of the robots in $\Gamma$ where at least one robot is at a pole and the other at a different circle or both robots are at antipodal positions in the same circle.

\begin{figure}[H]
    \centering
    \includegraphics[width=5in]{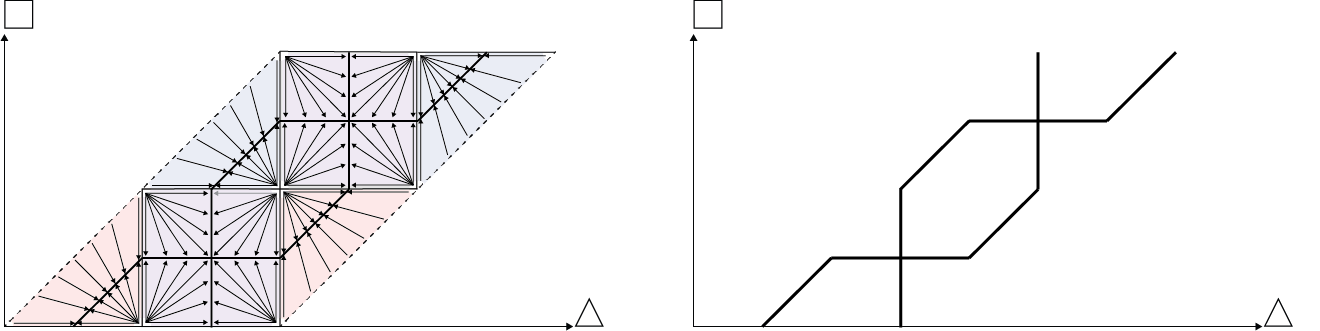}
    \caption{Homotopy traces of the retraction of $X$ into $Z$.}
    \label{spacextraces}
\end{figure}

We construct a retraction $r$ of the whole space $X$ into the spine $Z$. The homotopy deformation $H$ will follow the traces $H$($\textunderscore$,t) shown in figure \ref{spacextraces}.

\begin{figure}[H]
    \centering
    \includegraphics[width=2in]{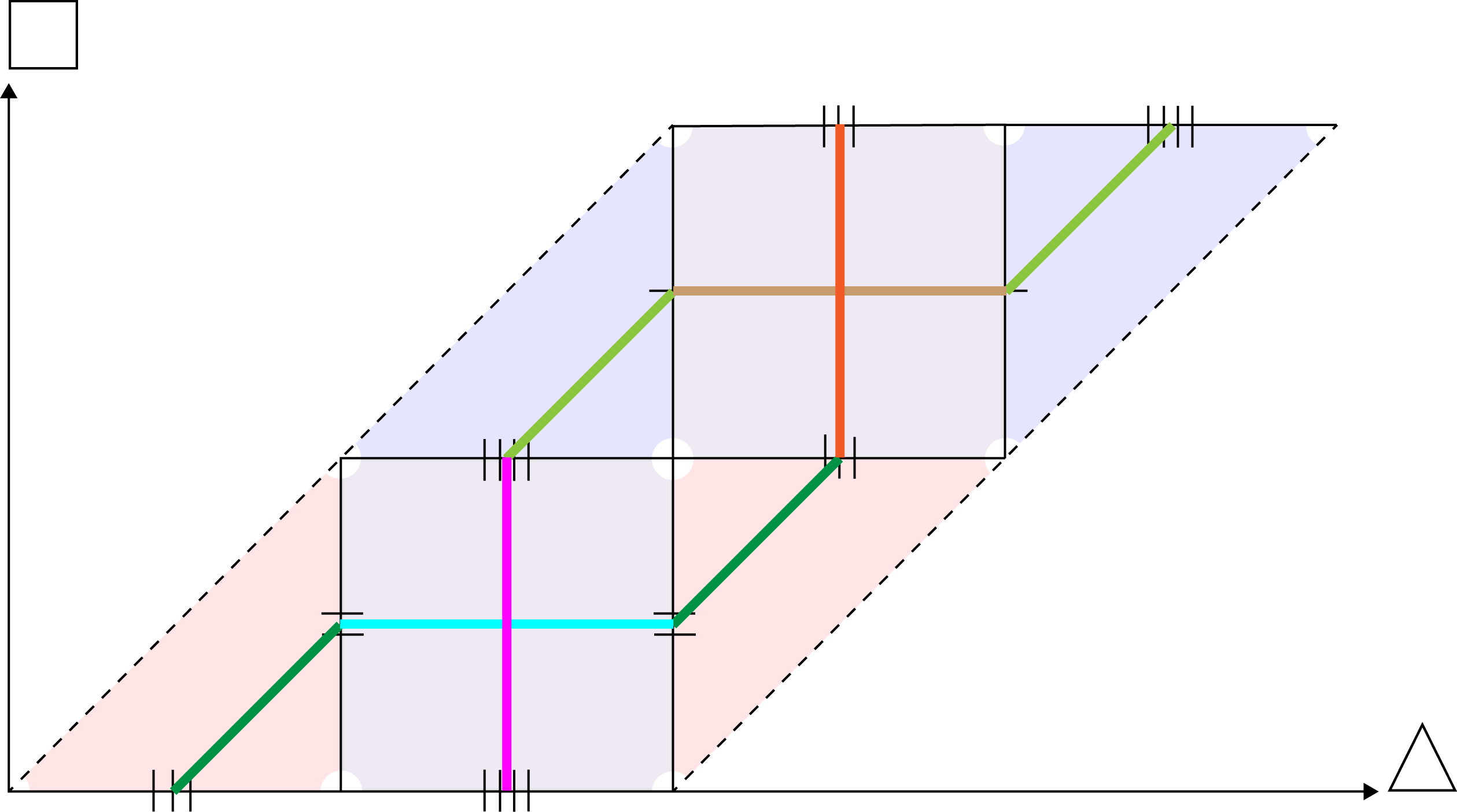}
    \caption{Spine $Z$.}
    \label{spacexspinecolor}
\end{figure}

The map $r$ is a deformation retraction since it is a retraction and the composition with the inclusion is homotopic to the identity map in $X$.We observe the following identifications of the segment extremes in $Z$ shown in figure \ref{spacexspinecolor}.

These identifications make the spine $Z$ homeomorphic to a chain of circles $C$. We color each segment to describe this homeomorphism.

\begin{figure}[H]
    \centering
    \includegraphics[width=4in]{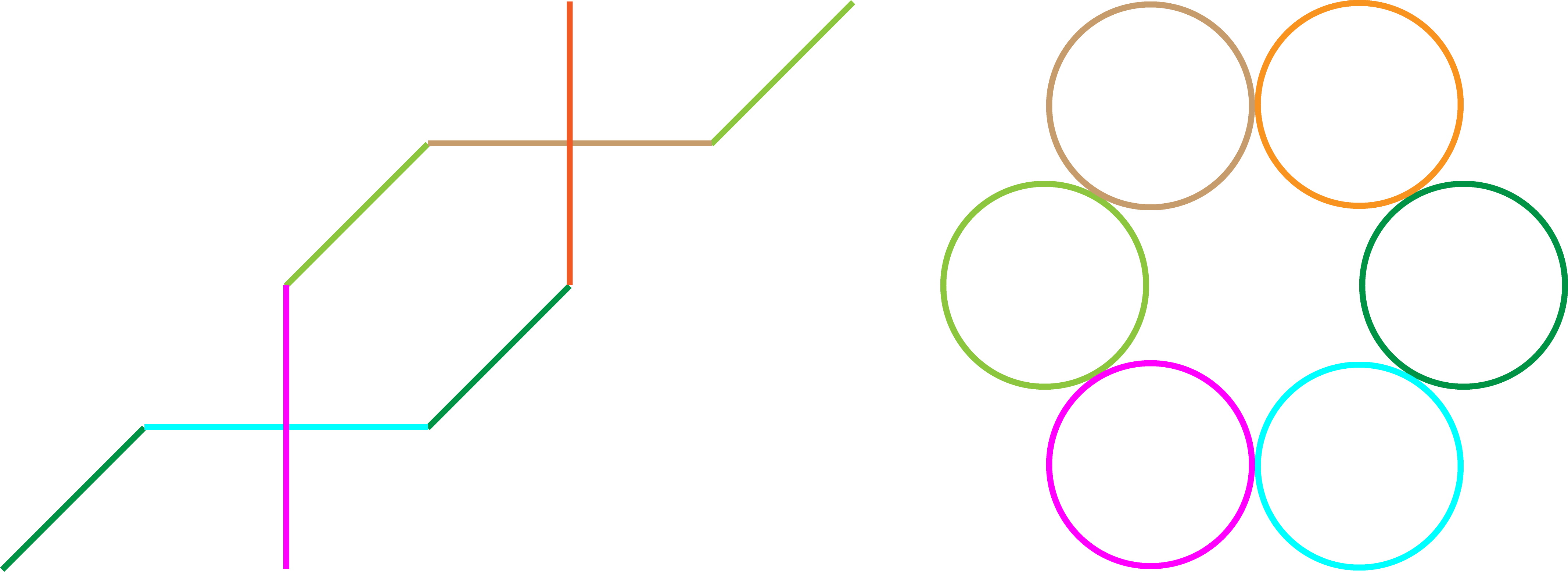}
    \caption{The spine $Z$ is homeomorphic to the chain $C$.}
    \label{spinecolor}
\end{figure}

\begin{figure}[H]
\centering
 \includegraphics[width=4in]{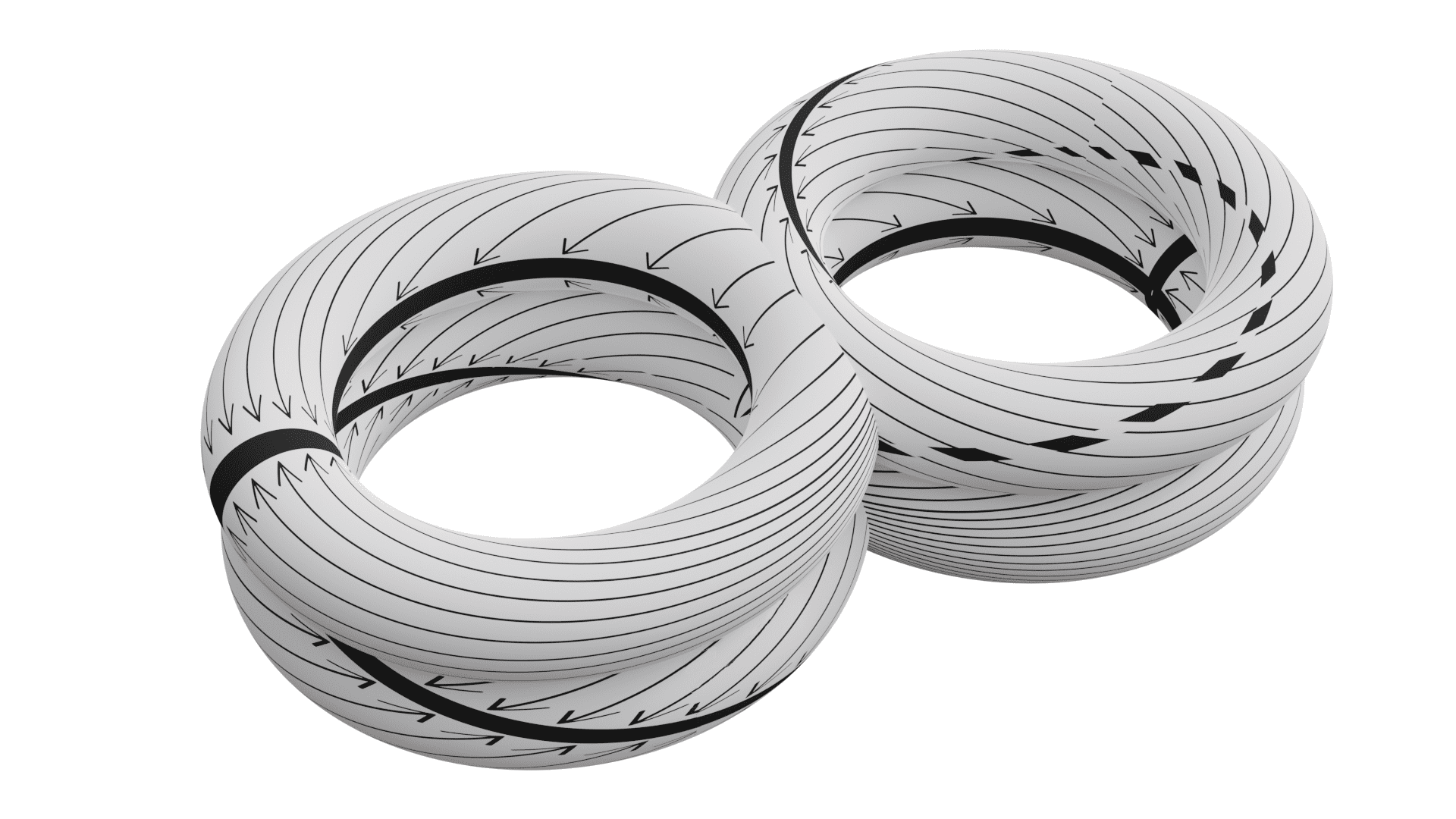}
\caption{Traces of the homotopy in $X$.}
\label{4tori3Dwirelarger}
 \end{figure}

\begin{figure}[H]
    \centering
    \includegraphics[width=4in]{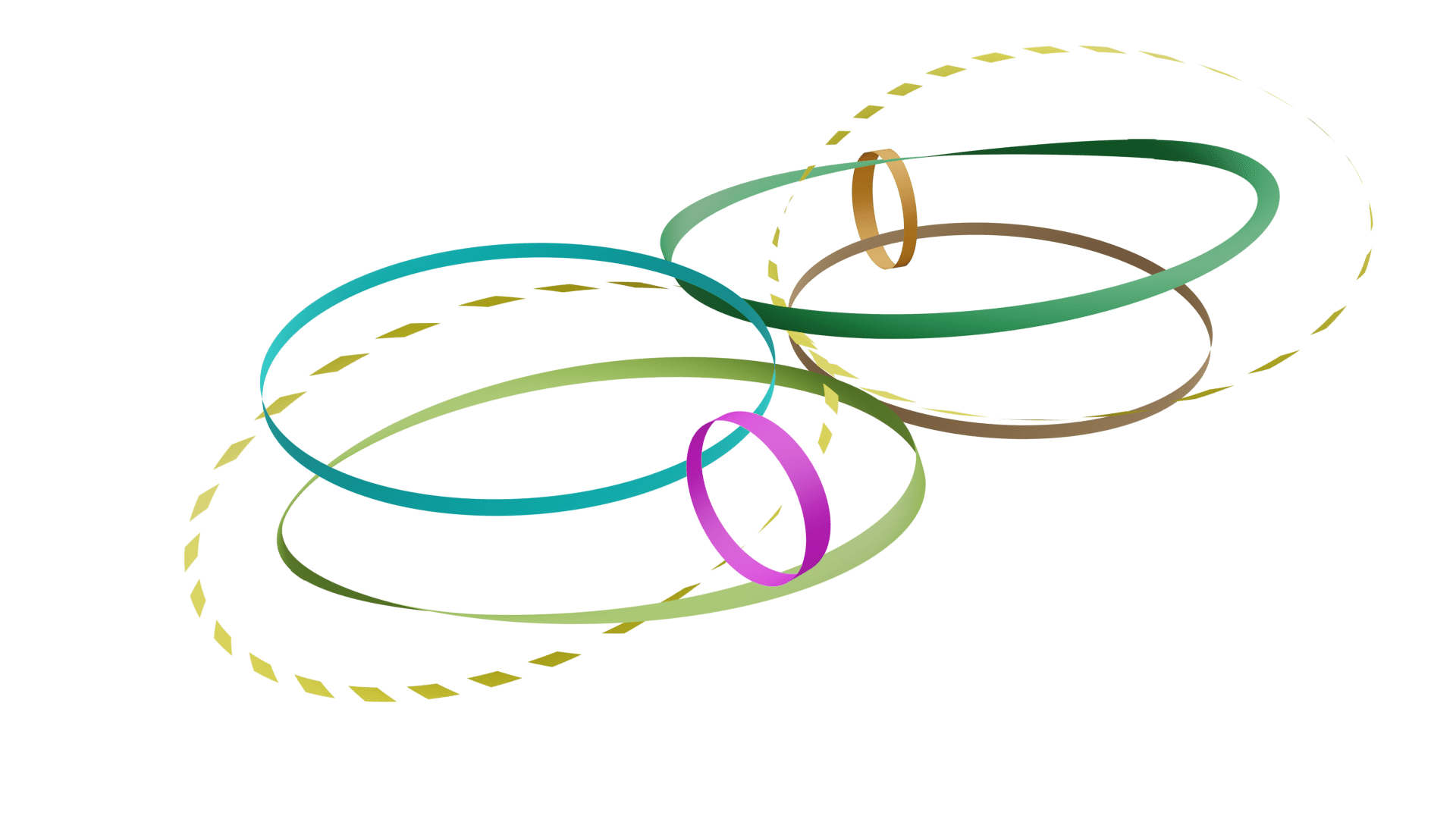}
    \caption{The chain $C$.}
    \label{spine3D}
\end{figure}

We can also visualize the chain of circles $C$ directly in the representation of $X$ embedded in $\mathbb R^3$. By following the traces of the homotopy, the two cylinders deform into circles and each torus minus a point deforms into a figure-eight. See figure \ref{4tori3Dwirelarger}. Note that in figures \ref{4tori3Dwirelarger} and \ref{spine3D}, the dashed lines are the diagonal $D$.

We can visualize in the following figure the chain of circles $C$ embedded  in the three-dimensional space.

\subsection{The wedge of seven circles}
\begin{figure}[H]
    \centering
    \includegraphics[width=4in]{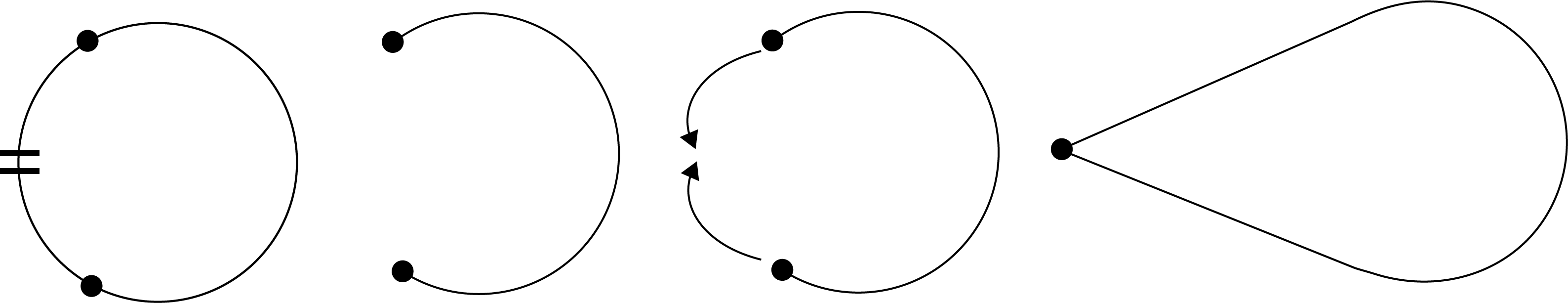}
    \caption{Contraction of a segment in a circle.}
    \label{ringcollapse}
\end{figure}

We will show in this section that the chain $C$  is homotopy equivalent to a wedge of seven circles $Y$. We 
perform a series of contractions in each circle as shown in figure \ref{ringcollapse}.

\begin{figure}[H]
    \centering
    \includegraphics[width=4in]{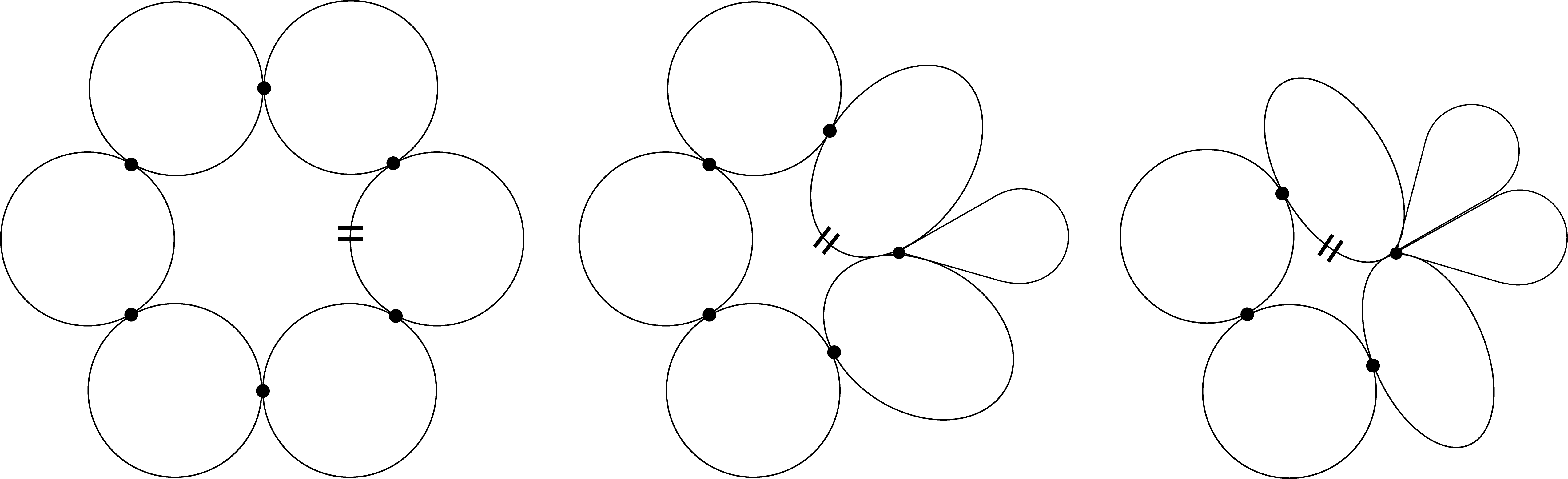}
    \caption{First stages of the deformation.}
    \label{chaincollapse1}
\end{figure}

\begin{figure}[H]
    \centering
    \includegraphics[width=4in]{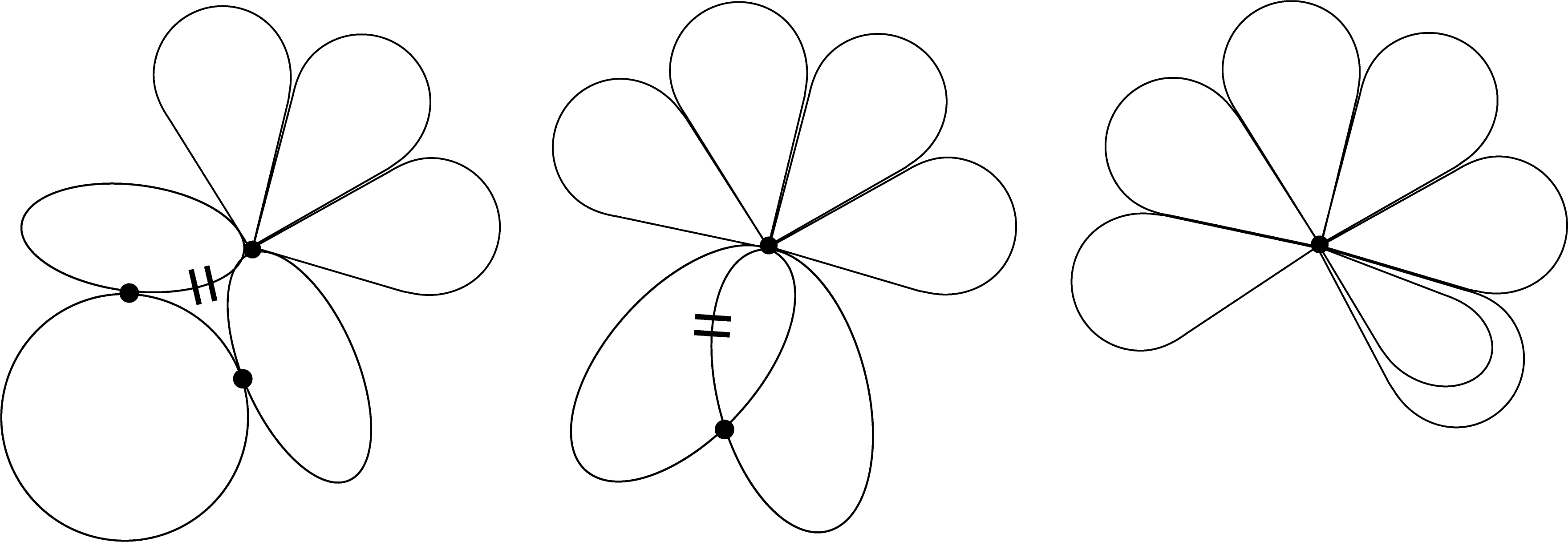}
    \caption{Last stages of the deformation.}
    \label{chaincollapse2}
\end{figure}

We repeat this deformation in each of the circles. See figures \ref{chaincollapse1} and  \ref{chaincollapse2}.

\begin{figure}[H]
    \centering
    \includegraphics[width=1in]{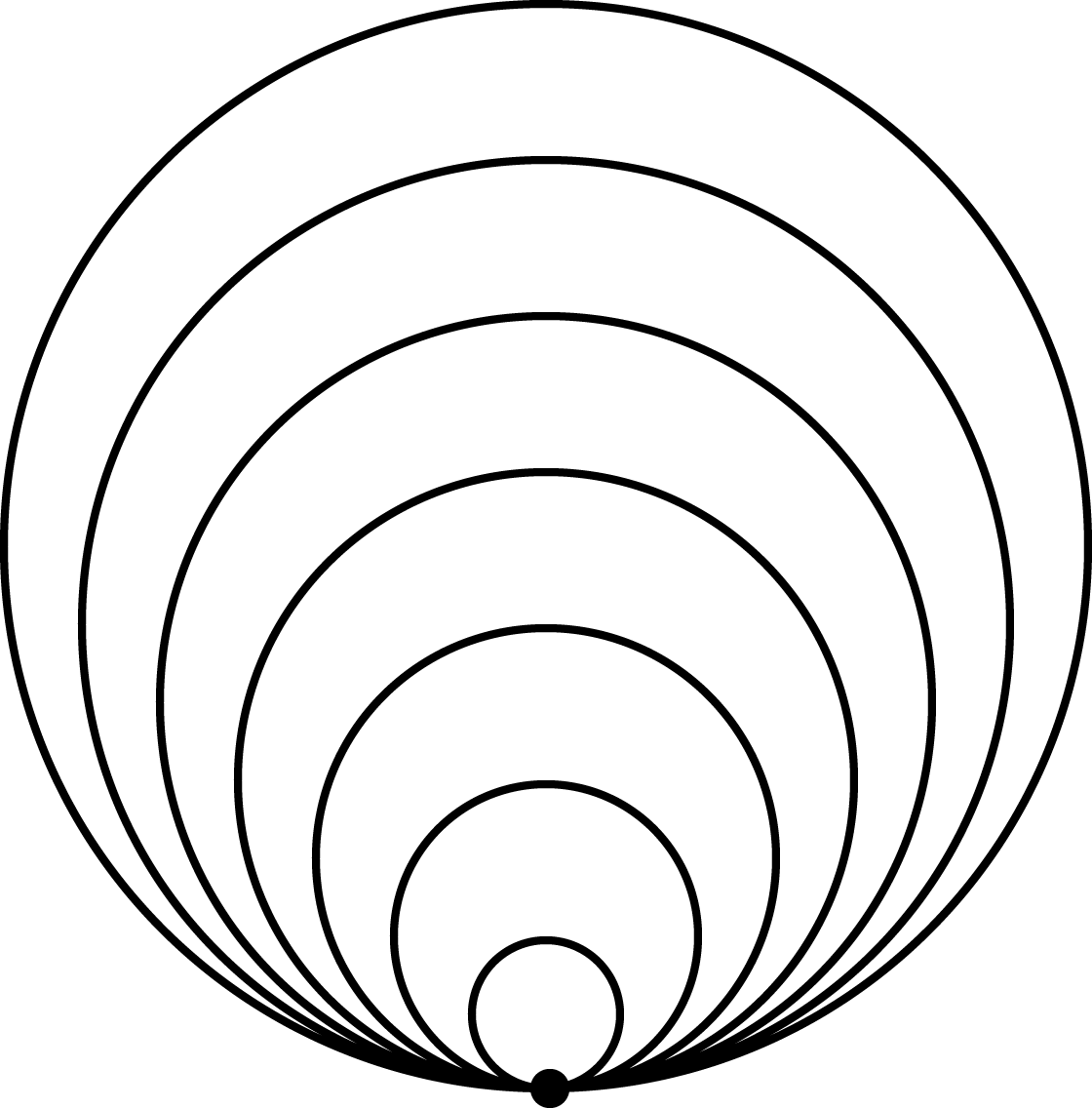}
    \caption{Wedge of seven circles $Y$. }
    \label{wedge1}
\end{figure}

The final space is a wedge of seven circles as shown in figure \ref{wedge1}.

\subsection{Calculation of the topological complexity of the configuration space}
The configuration space $X$ is homotopy equivalent to the spine $Z$ and $Z$ is homeomorphic to the chain of circles $C$, which is homotopy equivalent to the wedge $Y$, i.e.
    
    $$X \simeq Z \approx C \simeq Y.$$
    
Then, by Farber's theorem \ref{invariant} we have that 
    \begin{equation}\label{TC}
    TC(X) = TC(Z)=TC(C)=TC(Y).
   \end{equation}
Farber also studied and calculated the topological complexity of wedges of spheres:
   
            \begin{lemma}\cite{F3}\label{wedge}
    Let $W$ denote the wedge of $n$ spheres $S^m$, 
            $W = S^m \vee S^m \vee \cdots \vee S^m$. Then 
                $ TC(W) = \left\{
                        \begin{array}{ll}
                            2 & \mbox{if } n=1 \mbox{ and } m \mbox { is odd,} \\
                            3 & \mbox{if either } n > 1 \mbox{, or m is even.}
                        \end{array}
                    \right.
                    $
            \end{lemma}
                           For our space $Y$ we have that $m=1$ and $n=7$.    By lemma \ref{wedge}, we know that the topological complexity of a wedge of seven circles is $3$, then $TC(X)=3$ following equation \ref{TC}.
We know now that our algorithm will have to have at least $3$ continuous instructions.

\section{Algorithm in the configuration space}\label{Section6}
We will construct first our algorithm with three instructions in the spine $Z$ that is homotopy equivalent to $X$. Then we will extend the algorithm to the whole configuration space $X$ following the traces of the homotopy.

Recall that the spine $Z$ in the flat representation is a network consisting of two crosses and four sub-diagonals. 
   
\begin{figure}[H]
    \centering
    \includegraphics[width=3in]{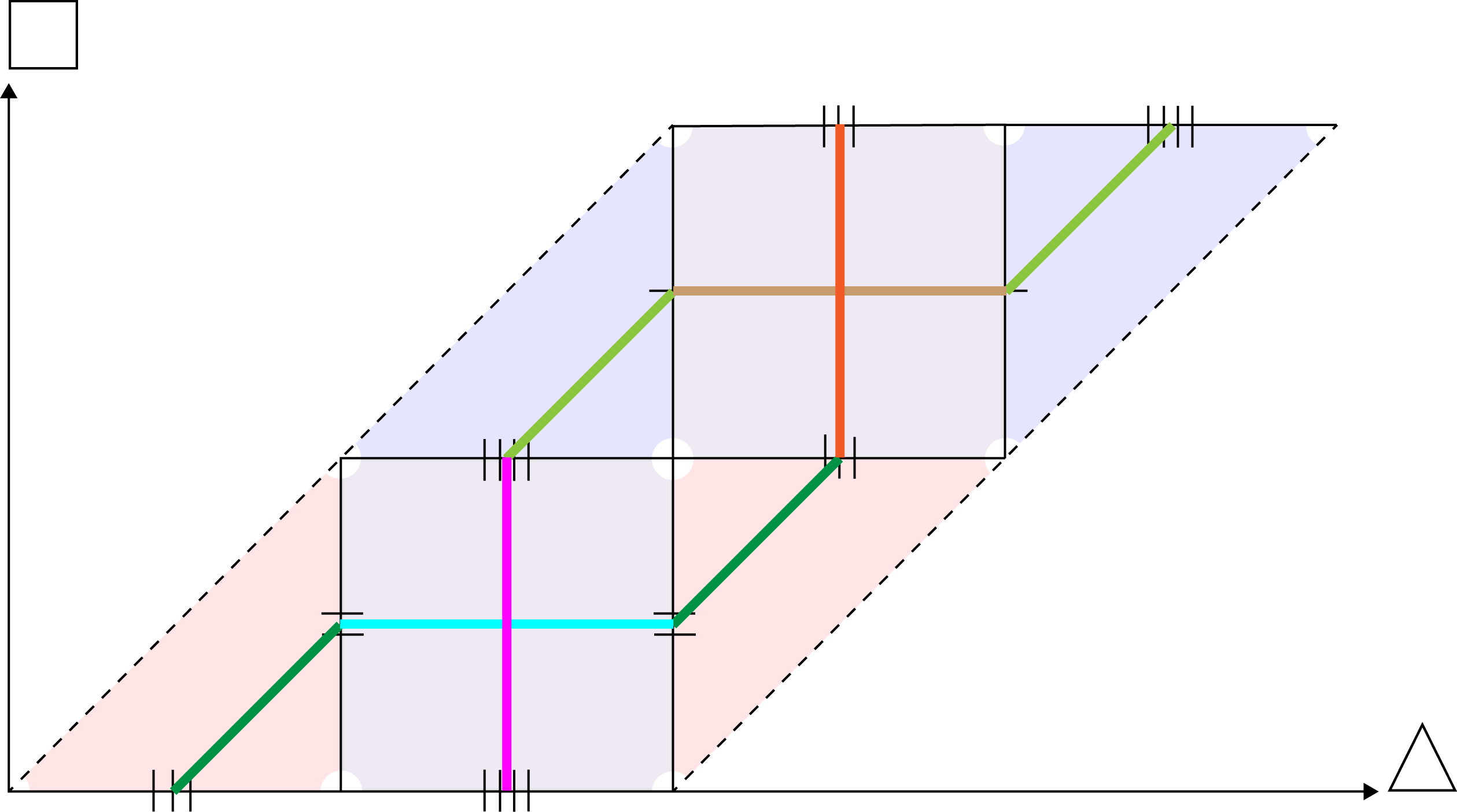}
    \caption{Space Z}
    \label{spacexthicker}
\end{figure}

When the state is in a diagonal segment in $Z$, we will say that the state is a {\em sub-diagonal state}. In the physical space $\Gamma$, this state will correspond to antipodal positions in the same circle. 
   
    \begin{figure}[H]
        \centering
        \includegraphics[width=4in]{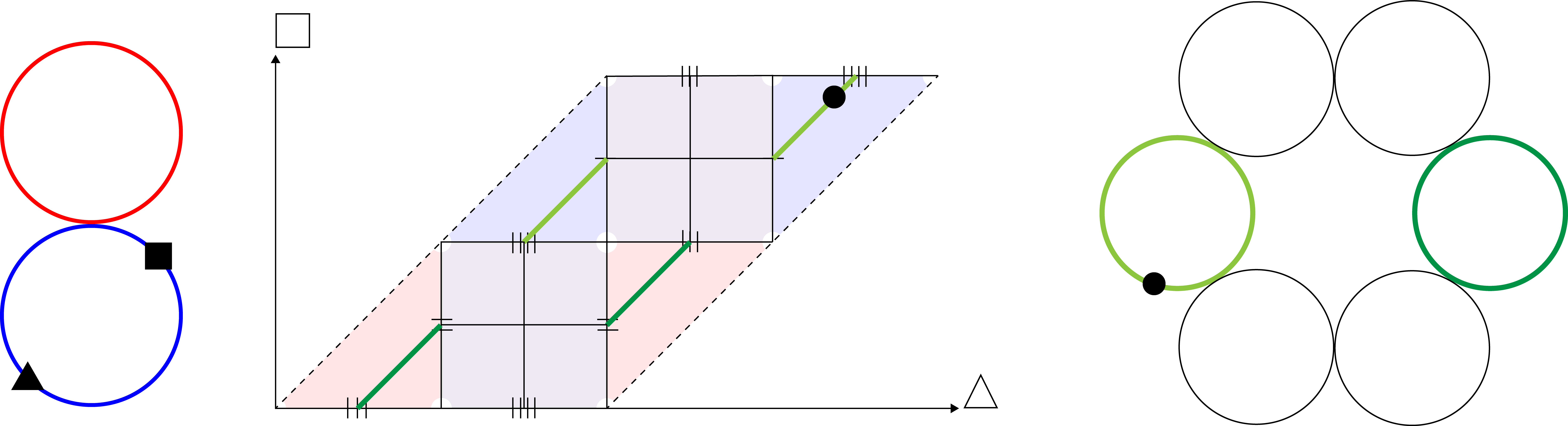}
        \caption{A sub-diagonal state.}
        \label{spinesubdiagonals}
    \end{figure}

         When a state is located in a cross-segment in $Z$, we will say that it is a {\em cross-state}. In the physical space $\Gamma$, cross-states translate as the position of at least one robot being at a pole and the other anywhere on the opposite circle.
         
    \begin{figure}[H]
        \centering
        \includegraphics[width=4in]{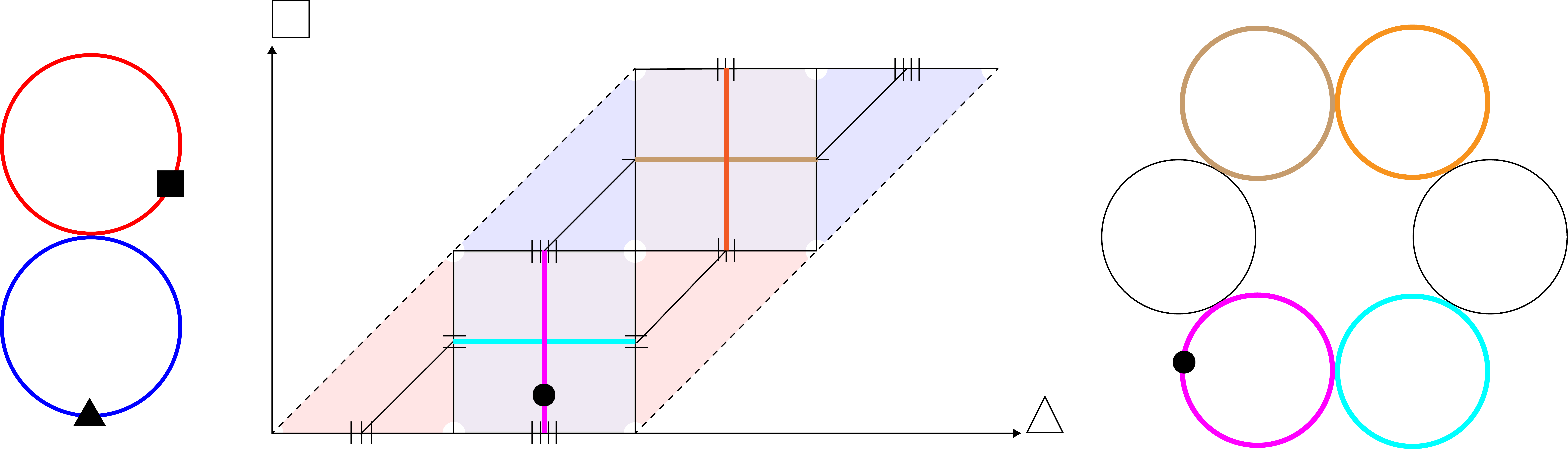}
        \caption{A cross-state.}
        \label{spinecross}
    \end{figure}
        
A {\em cross center} is a cross-state at the intersection of horizontal and vertical cross-segments. A cross center corresponds to the positions  in which both robots are at poles in different circles in $\Gamma$.
        
\begin{figure}[H]
    \centering
    \includegraphics[width=1in]{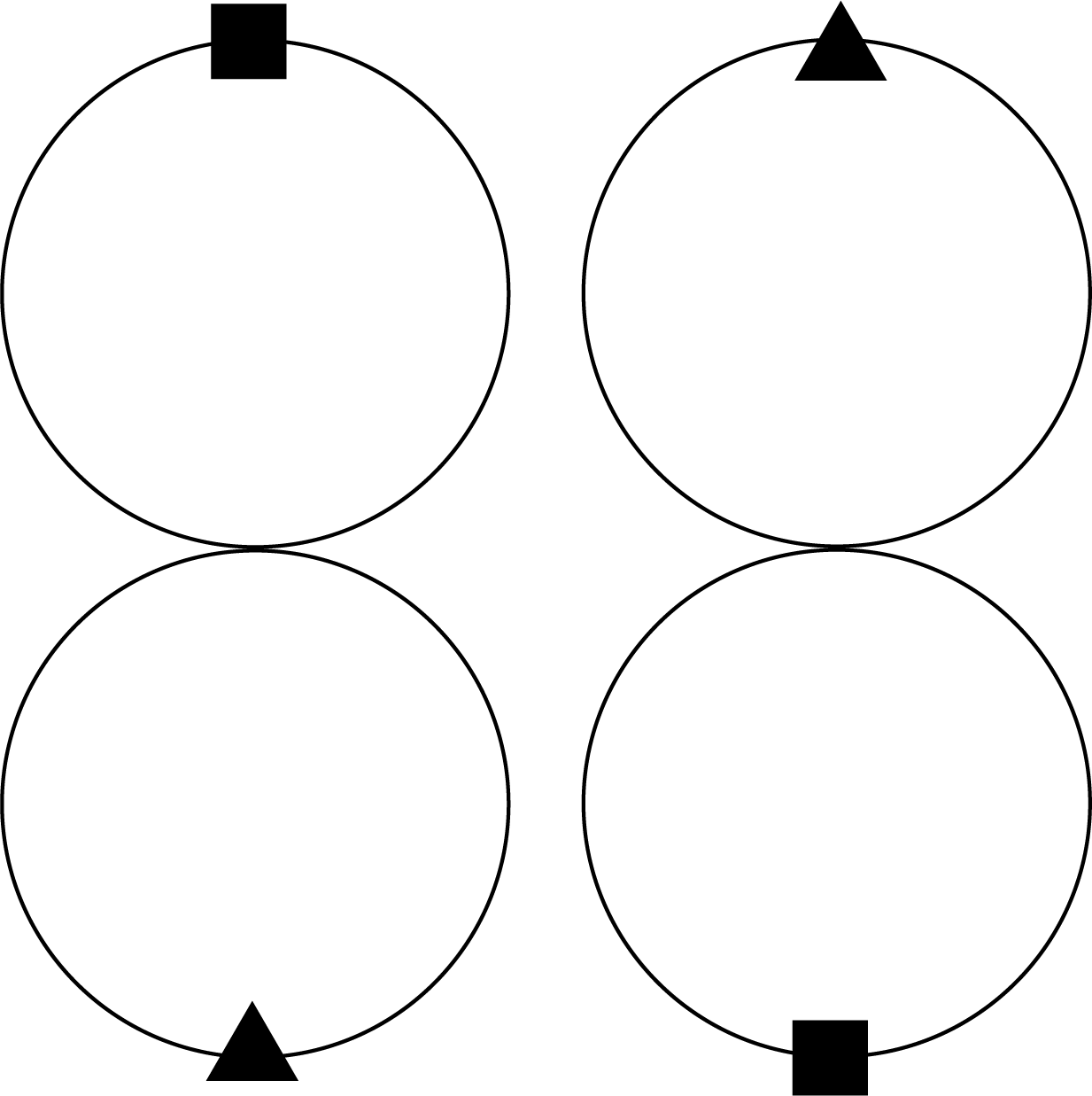}
    \caption{Cross centers in the physical space $\Gamma$.}
    \label{gammaVpoint}
\end{figure}

\begin{figure}[H]
    \centering
    \includegraphics[width=4in]{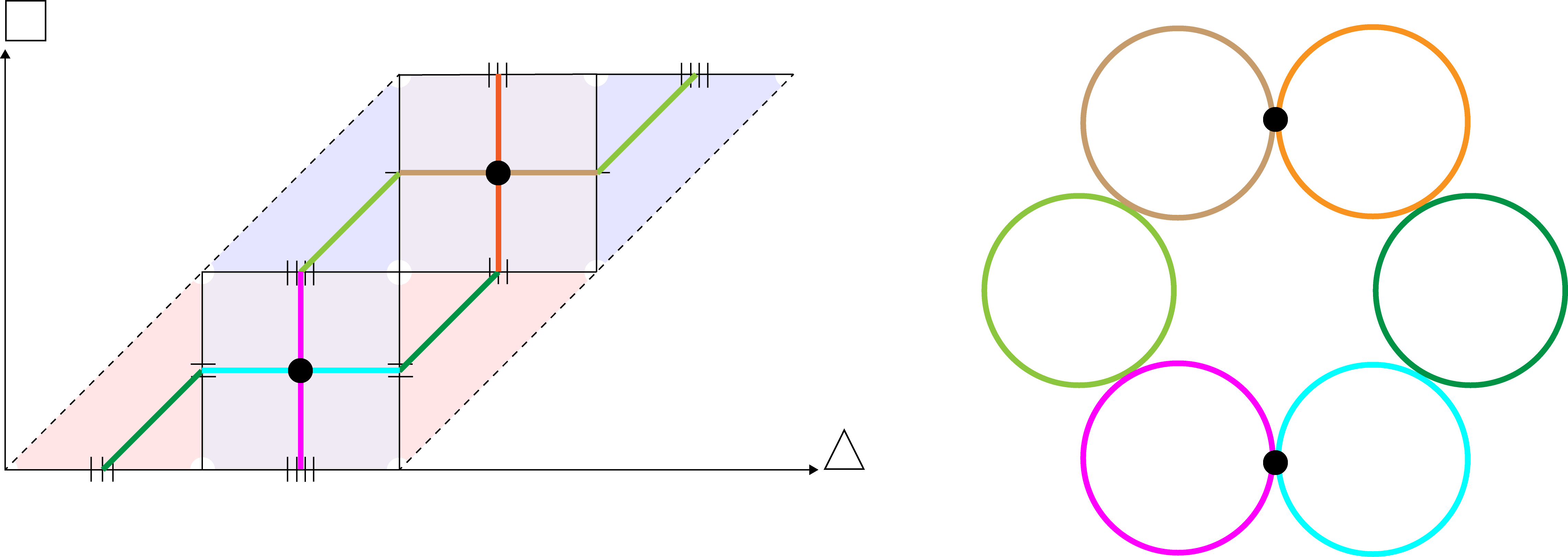}
    \caption{Cross centers in $Z$ and $C$.}
    \label{spineVpoint}
\end{figure}

        An {\em intersection point} is a state in $Z$ at the intersection between a cross-segment and a sub-diagonal segment.  In $\Gamma$, this state corresponds to positions in which one robot is at the center and the other at a pole. There are two types of intersection points: {\em $H$-points} located between a sub-diagonal segment and a horizontal cross-segment, and  {\em $V$-points} located between a sub-diagonal segment and a vertical cross-segment. An $H$-point corresponds in the physical space to a position in which the first robot is at the center and the second robot is at a pole. Conversely for $V$-points.

\begin{figure}[H]
    \centering
    \includegraphics[width=2in]{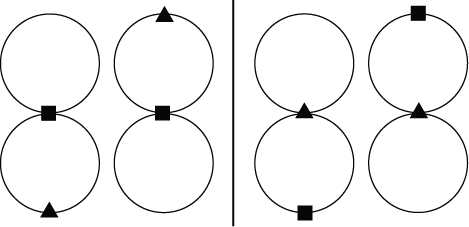}
    \caption{Intersection V-points(left) and H-points(right) in the physical space $\Gamma$.}
    \label{gammaHpoints}
\end{figure}
		
\begin{figure}[H]
    \centering
    \includegraphics[width=4in]{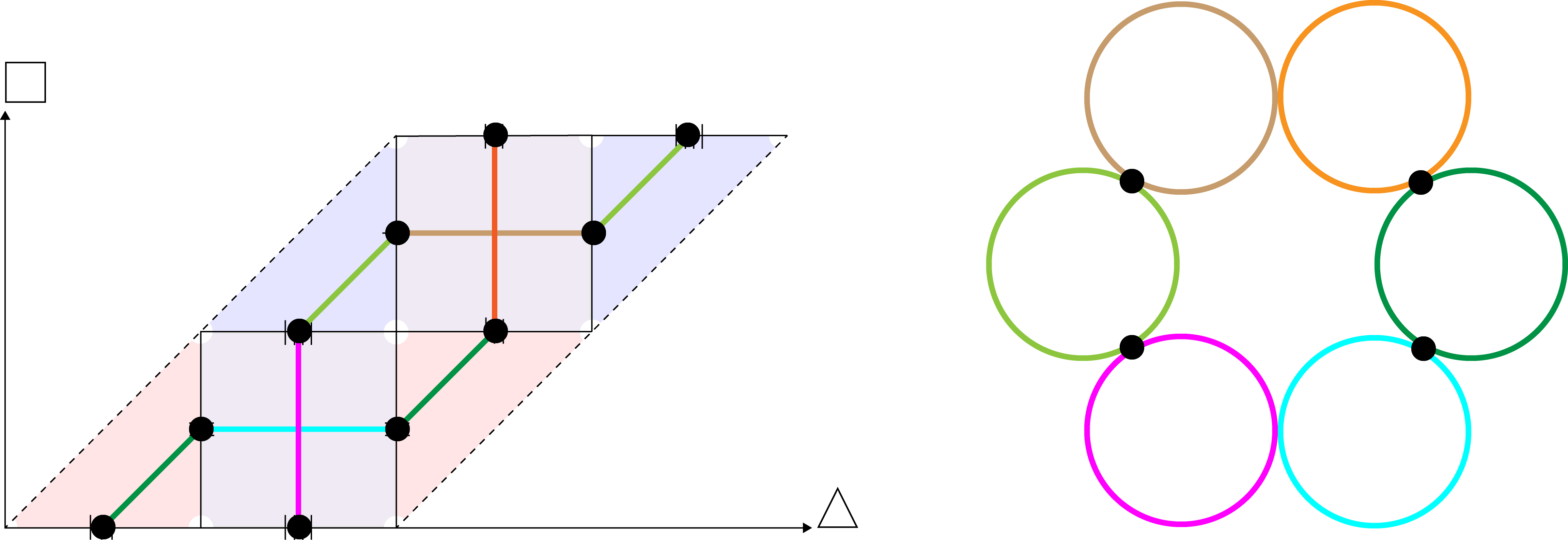}
    \caption{Intersection points in $Z$ and $C$.}
    \label{spineHpoints}
\end{figure}

The cross and sub-diagonal segments in $Z$ will constitute the circles in the chain $C$ whereas the intersection points and cross centers will be the vertices in $C$.

\subsection{Algorithm in the chain $C$}
The initial, final, and current states are the states corresponding to the initial, final, and current positions respectively.

If the current state is at a vertex in $C$, we will say that the {\em current circle} is the next immediate circle counterclockwise. Otherwise, the current circle is the circle where the current state is.
   
 Let $(I,F) = ((x_i, y_i ), (x_f,y_f))$ be a pair of  initial and final states in $C\times C$. We define the following subsets of $C\times C$:
\begin{itemize}
    \item $A = \left\{ (I,F) \in C \times C \ \lvert \ I \text{ is antipodal to } F \text{ in the same circle}\right\}$
    \item $V = \left\{ (I,F) \in C \times C \ \lvert \ I \text{ or } F \text{ is a vertex}\right\}$
    \item $V' = \left\{ (I,F) \in C \times C \ \lvert \ I \text{ and } F \text{ are vertices}\right\}$
\end{itemize} 

\begin{algorithm}\label{Algorithm in C}
Set current state to initial state $I$
\begin{description}
\item[Instruction 1]
For $(I,F) \in C \times C - (A \cup V)$:
    \begin{enumerate}
        \item Check if final state is in current circle.
        \begin{itemize}
            \item If true, take shortest path to final state.
            \item If false, move counterclockwise in current circle to next vertex and set current state to that vertex. Repeat 1.
        \end{itemize}
    \end{enumerate}
 \item[Instruction 2]   
For $(I,F) \in (A \cup V) - V'$:
    \begin{enumerate}
        \item[2.] Check if final state is in current circle.
        \begin{itemize}
            \item If true, check if final state is antipodal to current state.
            \begin{itemize}
                \item If true, go counterclockwise to final state.
                \item If false, take shortest path to final state.
            \end{itemize}
            \item If false, check if current state is a vertex.
            \begin{itemize}
                \item If true, move counterclockwise in current circle to next vertex and set current state to that vertex. Repeat 2.
                \item If false, take shortest path to next counterclockwise vertex and set current state to that vertex. Repeat 2.
            \end{itemize}
        \end{itemize}
    \end{enumerate}
    
    \item[Instruction 3]
For $(I,F) \in V'$:
    \begin{enumerate}
        \item[3.] Check if final state is next counterclockwise vertex.
        \begin{itemize}
            \item If true, go counterclockwise in the current circle to final state.
            \item If false, go counterclockwise in the current circle to next vertex and set current state to that vertex. Repeat 3.
        \end{itemize}
    \end{enumerate}
    \end{description}
\end{algorithm}
Each of these instructions is continuous in its domain.

\subsection{Algorithm in  $Z$}
By following the homeomorphism between $C$ and $Z$ we will translate now each of these instructions to the spine $Z$.

We define a distinguished direction of movement in $Z$ that will correspond to moving from vertices counterclockwise in the current circle in $C$.  The {\em positive direction in $Z$} is given by:
\begin{itemize}
\item from a cross center, move to the right in the horizontal segment;
\item from a $H$-point, move up in the diagonal segment;
\item from a $V$-point, move up in the vertical segment;
\item from any other point, move to the right in horizontal or diagonal segments and up in vertical segments.
\end{itemize}

\begin{figure}[H]
    \centering
    \includegraphics[width=5in]{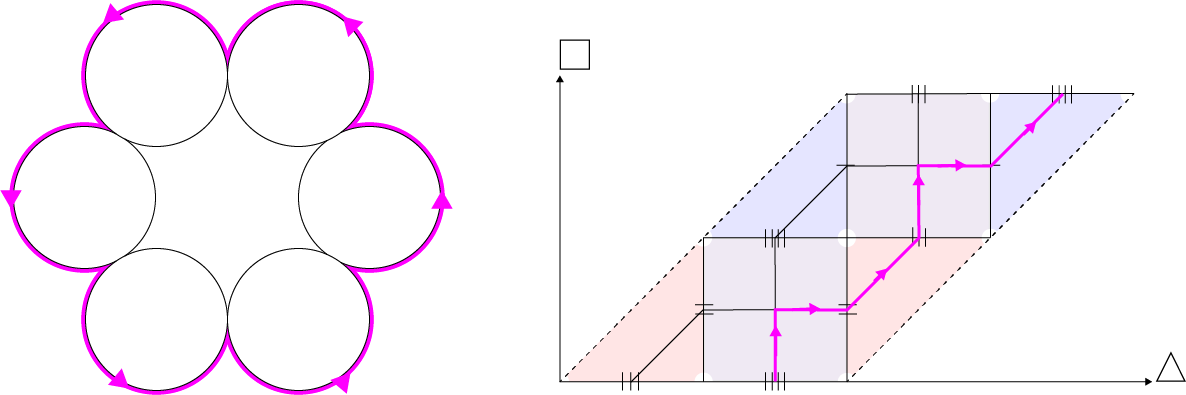}
    \caption{Positive direction in $C$ and $Z$.}
    \label{spinepositive}
\end{figure}

If the current state is at an intersection point or cross center in $Z$, we will say that the {\em current segment} is the next immediate segment in the positive direction. Otherwise, the current segment is where the current state is.

We consider $Z$ to be a metric graph where each edge has length one and the distance between two points is given by the infimum of the lengths of paths joining them.

Let $(I,F) = ((x_i, y_i ), (x_f,y_f))$ be a pair of  initial and final states in $Z\times Z$. We define the following subsets of $Z\times Z$:

\begin{itemize}
    \item $H = \{ (I,F) \in Z \times Z \ \lvert \  I \text{ and } F \text{ are on the same segment and } d(I,F)=\frac{1}{2} \}$
    \item $J = \{ (I,F) \in Z \times Z \ \lvert \ I \text{ or } F \text{ are at a cross center or intersection point}\}$
    \item $J' = \{ (I,F) \in Z \times Z \ \lvert \ I \text{ and } F \text{ are both at cross centers or intersection points}\}$
\end{itemize} 

\begin{algorithm}\label{Algorithm in Z}
Set current state to initial state $I$

\begin{description}
\item[Instruction 1]
For $(I,F) \in Z \times Z - (H \cup J)$:
    \begin{enumerate}
        \item Check if final state is in current segment.
        \begin{itemize}
            \item If true, take shortest path to final state.
            \item If false, move in the positive direction in current segment to next point of intersection and set current state to that point of intersection. Repeat 1.
        \end{itemize}
            \end{enumerate}
\item[Instruction 2]
For $(I,F) \in (H \cup J) - J'$:
    \begin{enumerate}
        \item[2.] Check if final state is in current segment.
        \begin{itemize}
            \item If true, check if the distance from current state to final state is $\frac{1}{2}$.
            \begin{itemize}
                \item If true, move in the positive direction to final state.
                \item If false, take shortest path to final state.
            \end{itemize}
            \item If false, check if current state is a point of intersection.
            \begin{itemize}
                \item If true, move in the positive direction along current segment to next point of intersection and set current state to that point of intersection. Repeat 2.
                \item If false, take shortest path, in the positive direction, to next point of intersection and set current state to that point of intersection. Repeat 2.
            \end{itemize}
        \end{itemize}
    \end{enumerate}
\item[Instruction 3]
For $(I,F) \in J'$:
    \begin{enumerate}
        \item[3.] Check if final state is the next point of intersection in the positive direction.
        \begin{itemize}
            \item If true, move in the positive direction along the current segment to final state.
            \item If false, move in the positive direction along the current segment to next point of intersection and set current state to that point of intersection. Repeat 3.
        \end{itemize}
    \end{enumerate}
    \end{description}
\end{algorithm}

\subsection{Algorithm in $X$}
We will extend now the algorithm \ref{Algorithm in Z} to the whole configuration space.

Given an initial state $I$,  let $I_s$ be the state corresponding to the point of intersection between the trace of the homotopy passing through $I$ and the spine $Z$. Analogously, $F_s$ is the corresponding state to the final one in the spine.

\begin{algorithm}\label{Algorithm in X}
Let $I$ and $F$ be the initial and final states. 
\begin{enumerate}
    \item Find $(I_s,F_s)$.
    \item\label{1} Move state $I$ to $I_s$.
    \item \label{2} Execute algorithm \ref{Algorithm in Z} for $(I_s,F_s)$.
    \item \label{3} Move state $F_s$ to $F$.
\end{enumerate}
\end{algorithm}
Given the initial and final states shown in figure \ref{move1}, the following figures illustrate algorithm \ref{Algorithm in X} for this case.

\begin{figure}[H]
    \centering
    \includegraphics[width=3in]{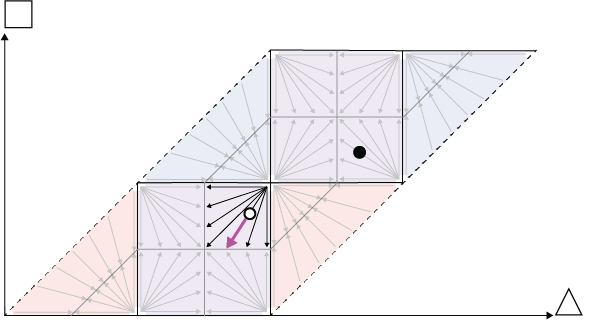}
    \caption{Step \ref{1}.}
    \label{move1}
\end{figure}

\begin{figure}[H]
    \centering
    \includegraphics[width=3in]{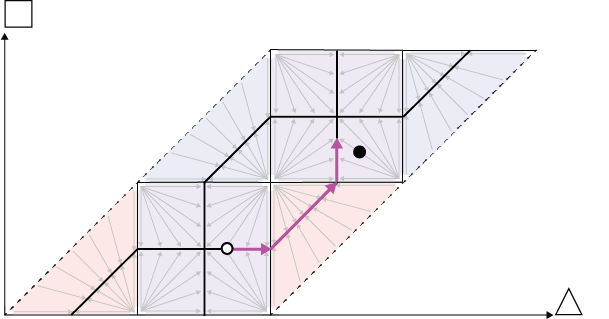}
     \caption{Step \ref{2}.}
    \label{move2}
\end{figure}

\begin{figure}[H]
    \centering
    \includegraphics[width=3in]{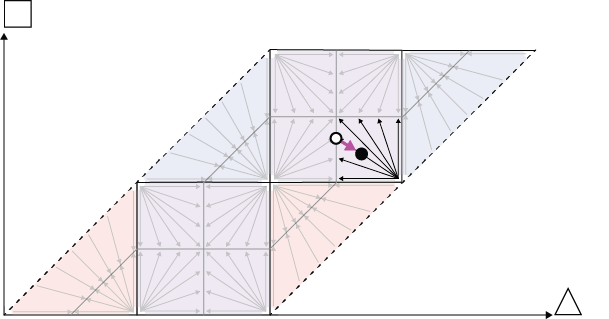}
     \caption{Step \ref{3}.}
    \label{move3}
\end{figure}

\section{Algorithm in the physical space}\label{Section7}
We will translate now our algorithm \ref{Algorithm in X} to the physical space $\Gamma$.

Observe that states in the spine $Z$ correspond to positions in which at least one robot is at a
pole if the robots are in different circles and antipodal positions if the robots are in the same circle.

\subsection{States $(I_s,F_s)$ in the physical space $\Gamma$}
To translate the first step of the algorithm, we need to find the positions of the robots in  $\Gamma$ corresponding  to the intersection of the traces with the spine.

\begin{figure}[H]
    \centering
    \includegraphics[width=3in]{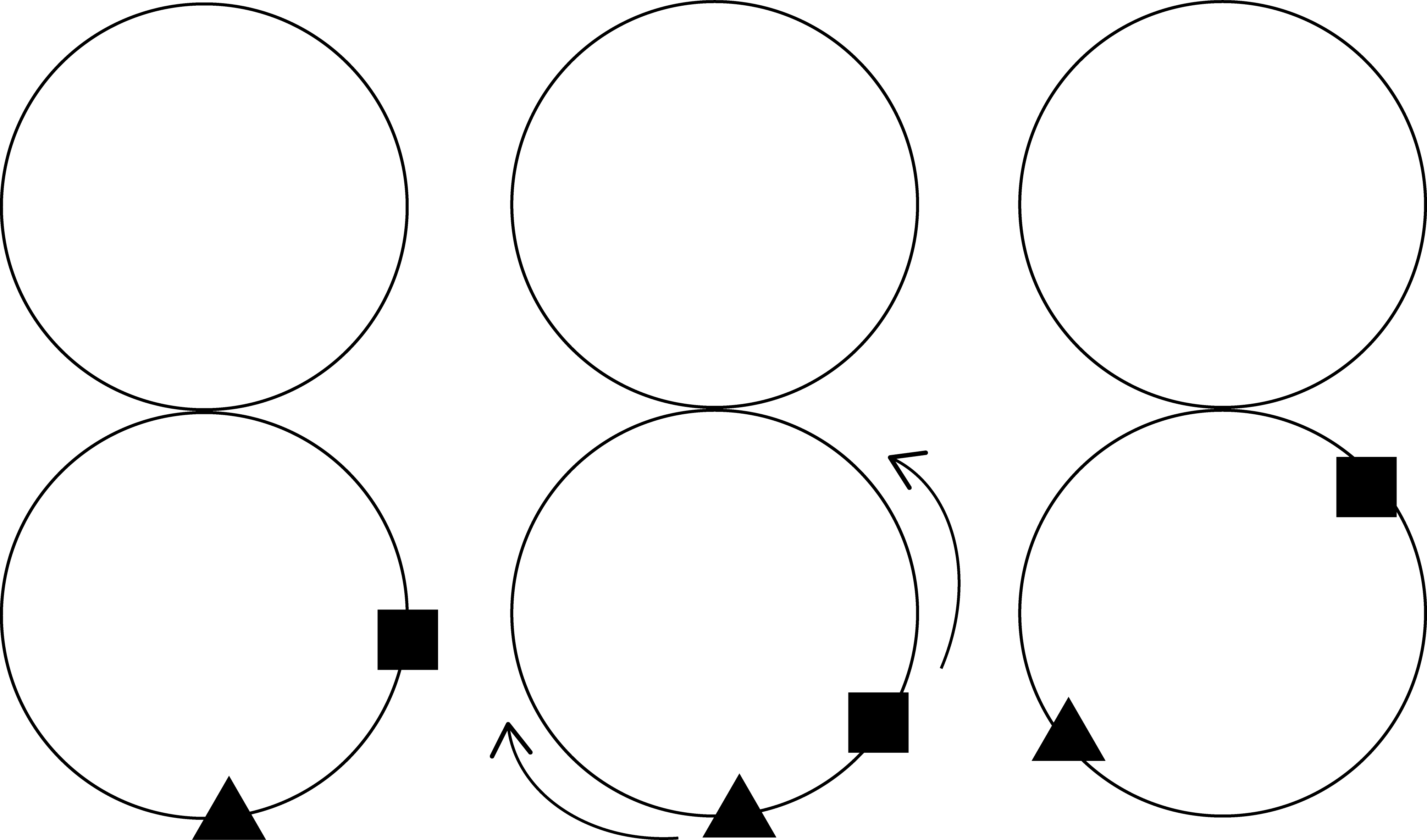}
    \caption{Moving to antipodal positions}
    \label{samecircle}
\end{figure}

\begin{figure}[H]
    \centering
    \includegraphics[width=3in]{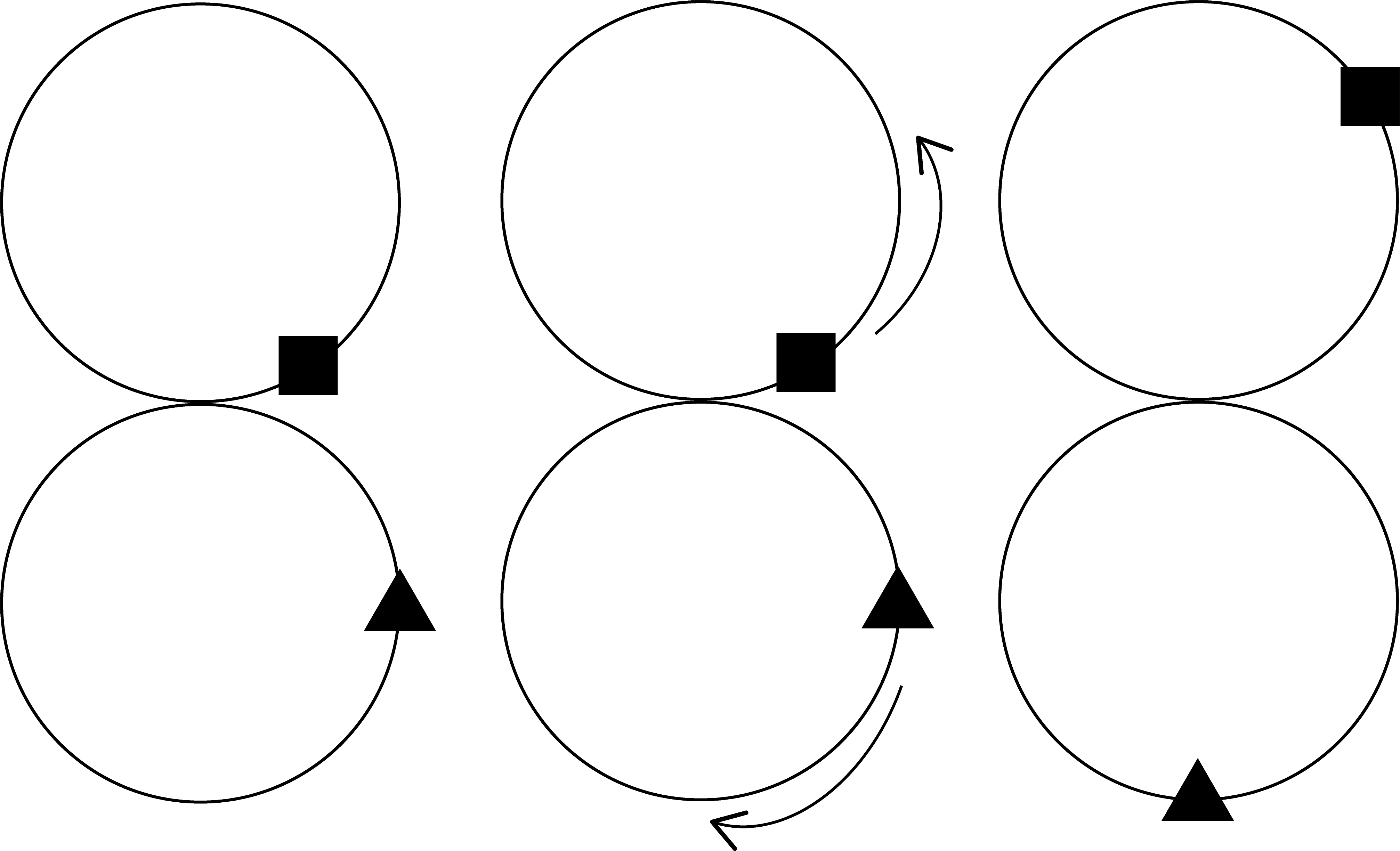}
    \caption{Moving until one robot reaches the pole.}
    \label{differentcircle}
\end{figure}

\begin{figure}[H]
    \centering
    \includegraphics[width=3in]{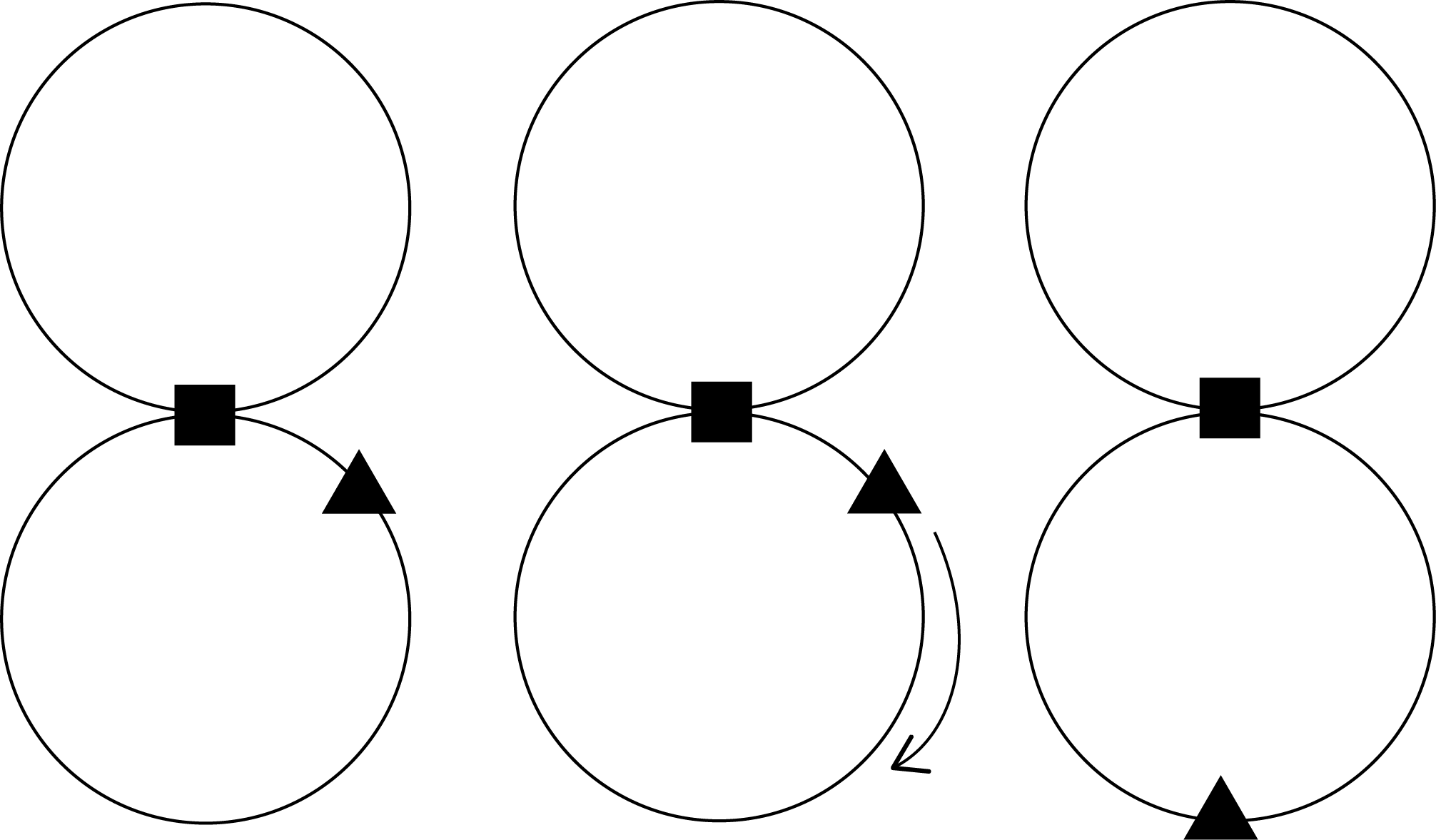}
    \caption{Robot in center scenario.}
    \label{special}
\end{figure}

If the robots are in the same circle, move the robots away from each other until they are in antipodal positions. See figure \ref{samecircle}. If they are in different circles, move them away from the center until at least one of them reaches a pole position. See figure \ref{differentcircle}. If one robot is at the center, move the other robot until it reaches a pole position. See figure \ref{special}.

The ratio between the speeds at which the robots are moving is determined by the slope of the traces of the homotopy discussed in section \ref{traces}. If a trace of the homotopy is given by the curve $(x(t), y(t))$, then the slope of the trace is given by the ratio between the speeds $y'(t)$ and $x'(t)$.

\subsection{Algorithm in the physical space for the positions in $Z$}

We will translate here the movement in the positive direction in $Z$ to the physical space.  Recall that a cross-center state corresponds to both robots being at poles, and intersection points in $Z$ correspond to one robot being at a pole and the other at the center. In the physical space, we will call these positions {\em vertex positions}.

The {\em positive direction in $\Gamma$} is given by:
\begin{itemize}
\item if both robots are at poles, move robot $1$ counterclockwise;
\item if the first robot is at the center and the other at a pole, move both counterclockwise;
\item if the first robot is at a pole and the other at the center, move the second robot  counterclockwise.
\end{itemize}

\begin{figure}[H]
    \centering
    \includegraphics[width=3in]{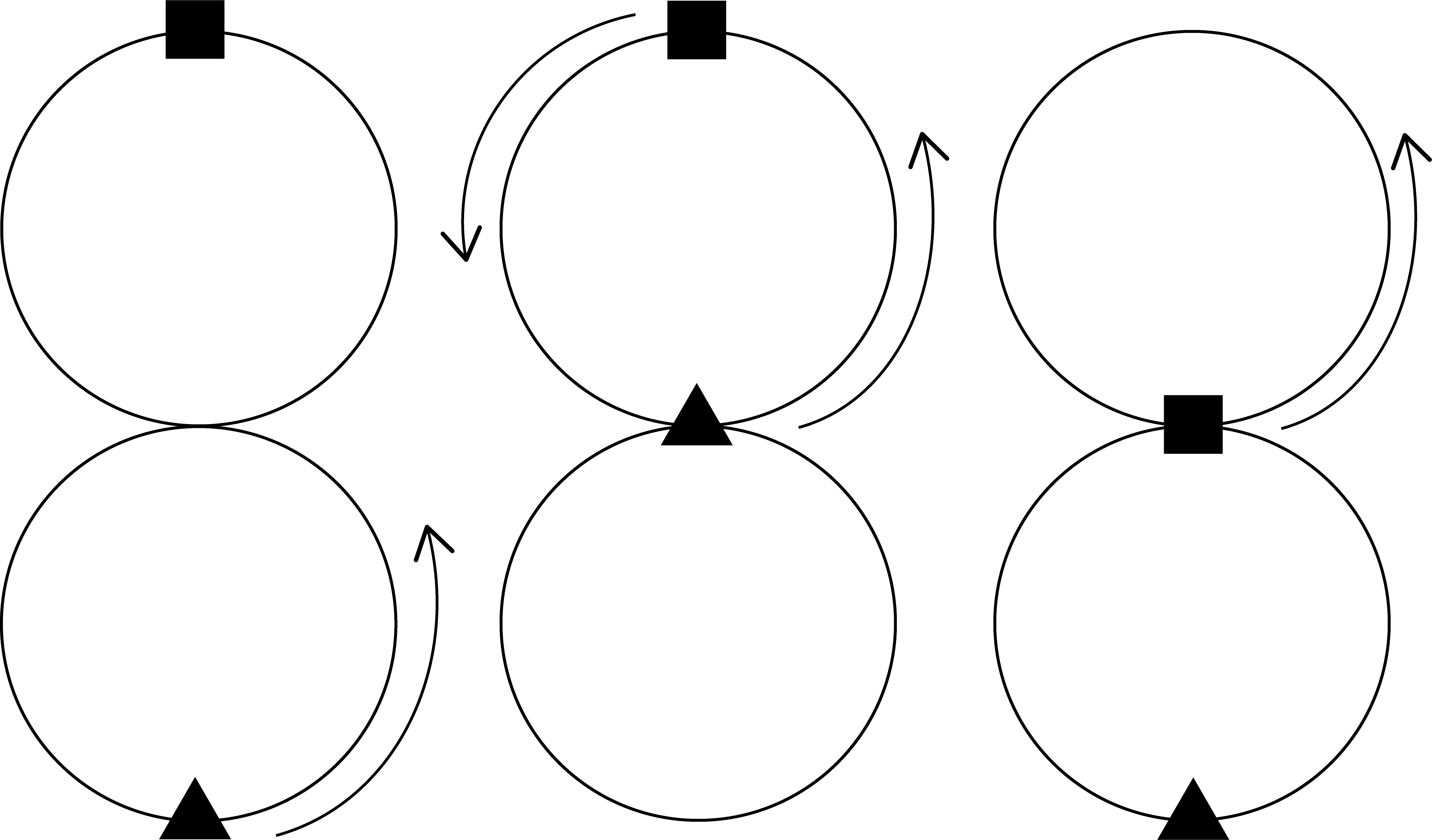}
    \caption{Positive direction in $\Gamma$.}
    \label{gammapositive}
\end{figure}

If the current position is a vertex position, we will say that its {\em neighborhood} is the set of positions determined by moving the current position in the positive direction until it circles back to itself. Otherwise, the neighborhood is given by moving counterclockwise the robot that is not at a pole all around its circle if the robots are at different circles or moving counterclockwise both robots together if they are antipodal in the same circle. 

\begin{figure}[H]
    \centering
    \includegraphics[width=5in]{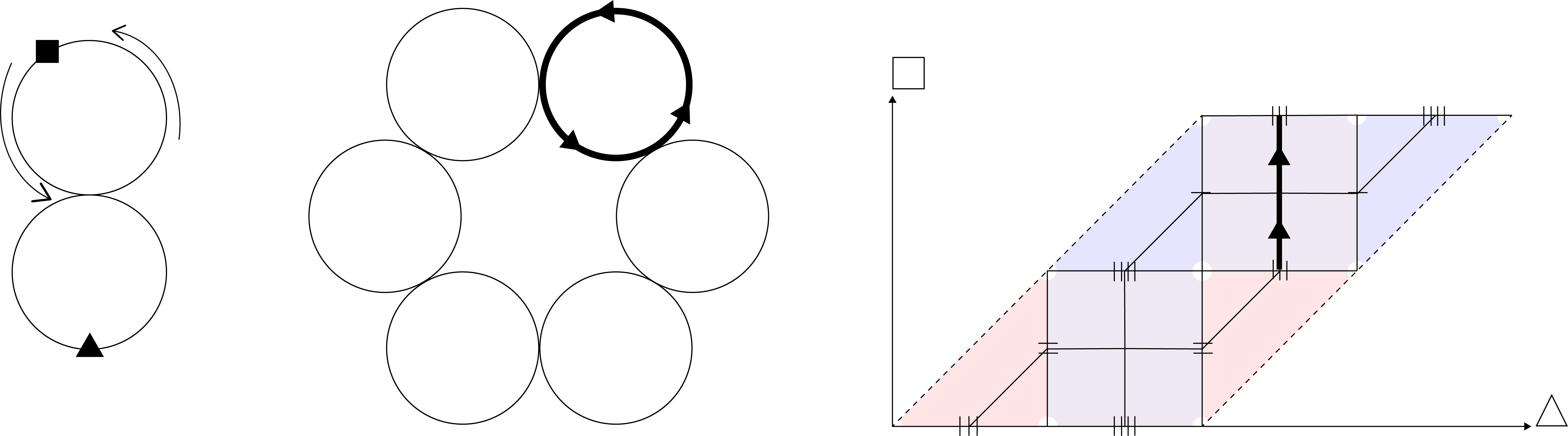}
    \caption{A neighborhood of a non-vertex position.}
    \label{diagonalneighborhood}
\end{figure}

\begin{figure}[H]
    \centering
    \includegraphics[width=5in]{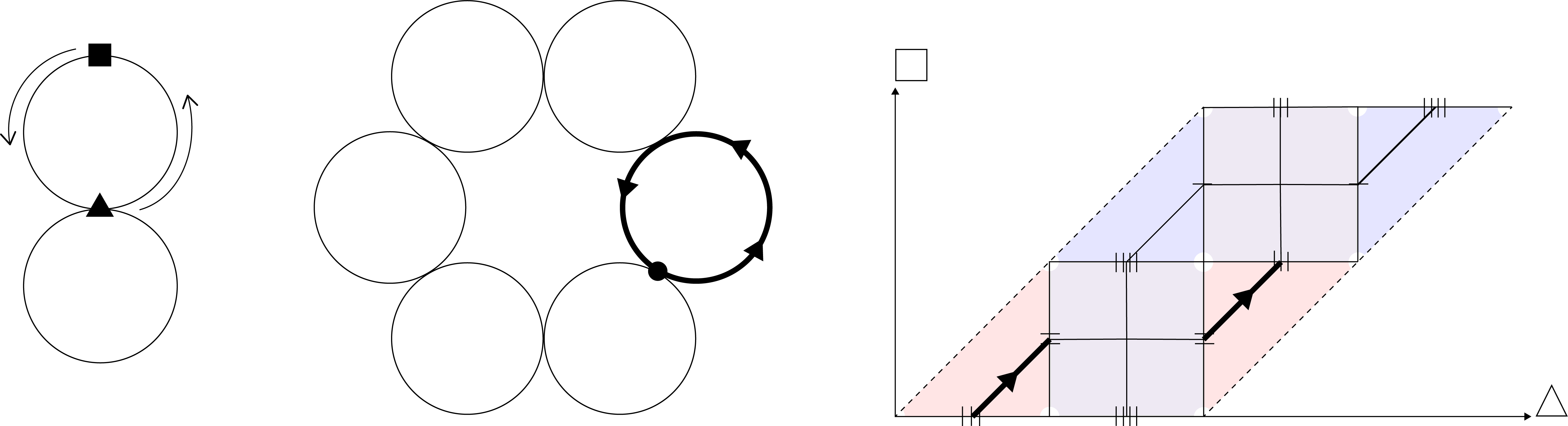}
    \caption{A neighborhood of a vertex position.}
    \label{crossneighborhood}
\end{figure}

Let $(I,F) = ((x_i, y_i ), (x_f,y_f))$ be a pair of  initial and final positions in $\Gamma\times \Gamma$. We define the following subsets of $\Gamma\times \Gamma$:

\begin{itemize}
    \item $K = \{ ((x_i, y_i ), (x_f,y_f)) \ \lvert  \ x_i \text{ and } x_f \text{ are antipodal} 
    \text{ or } y_i \text{ and } y_f \text{ are antipodal}\}$
    \item $L = \{((x_i, y_i ), (x_f,y_f)) \ \lvert \ (x_i, y_i ) \text{ or } (x_f,y_f) \text{ is a vertex position}  \}$
    \item $L' = \{((x_i, y_i ), (x_f,y_f)) \ \lvert \ (x_i, y_i ) \text{ and } (x_f,y_f) \text{ are vertex positions}\}$
\end{itemize}

\begin{algorithm}\label{Algorithm in the physical space for positions in Z}
Set current position to initial position $I$.
\begin{description}
\item[Instruction 1]
For $(I,F) \in \Gamma\times \Gamma - (K \cup L)$:
    \begin{enumerate}
        \item Check if final position is in the neighborhood of the current position.
        \begin{itemize}
            \item If true, move each robot simultaneously to their final position taking the shortest path.
            \item If false, move in the positive direction in current neighborhood until next vertex position. Set current position to this one. Repeat 1.
\end{itemize}
\end{enumerate}
    \item[Instruction 2]
    For $(I,F) \in (K \cup L) - L'$:
    \begin{enumerate}
        \item[2.] Check if final position is in the current neighborhood.
        \begin{itemize}
            \item If true, check if at least one of the robots have its initial and final positions antipodal.            
            \begin{itemize}
                \item If true, move in the positive direction to final position.
                \item If false, move each robot simultaneously to their final position taking the shortest path.
            \end{itemize}
            \item If false, check if current position is a vertex position.
                        \begin{itemize}
                \item If true, move in the positive direction in current neighborhood to the next vertex. Set current position to this one. Repeat 2.
                   \item If false, take shortest path to next vertex position in the positive direction and set current position to this one. Repeat 2.
            \end{itemize}
        \end{itemize}
    \end{enumerate}
\item[Instruction 3]
For $(I,F) \in L'$:
    \begin{enumerate}
        \item[3.] Check if final position is the next vertex position in the positive direction.
        \begin{itemize}
            \item If true, move in the positive direction in the current neighborhood to final position.
            \item If false, move in the positive direction in the current neighborhood to next vertex position and set current state to that position. Repeat 3.
        \end{itemize}
    \end{enumerate}
    \end{description}
\end{algorithm}

\subsection{Complete algorithm in the physical space $\Gamma$}

We will write now the complete algorithm for all positions in $\Gamma$.

\begin{algorithm}
Let $I$ and $F$ be the initial and final positions. 
\begin{enumerate}

    \item If the robots are in the same circle, move the robots away from each other until they are in antipodal positions.

    \item If they are in different circles, move them away from the center until at least one of them reaches a pole position. 

    \item Repeat steps $1$ and $2$ with the final position. Let us call initial
$Z$-position and final $Z$-position the output of this step.

  \item Execute algorithm \ref{Algorithm in the physical space for positions in Z} for the initial
$Z$-position and final $Z$-position.
    \item Move back from the final $Z$-position to the final position reversing the movement done in the first step.
\end{enumerate}

\end{algorithm}

In the following case scenarios, we describe the movements in the physical space as well as in the configuration space.

\subsubsection{Case scenario 1 }
\begin{figure}[H]
    \centering
    \includegraphics[height=1.5in]{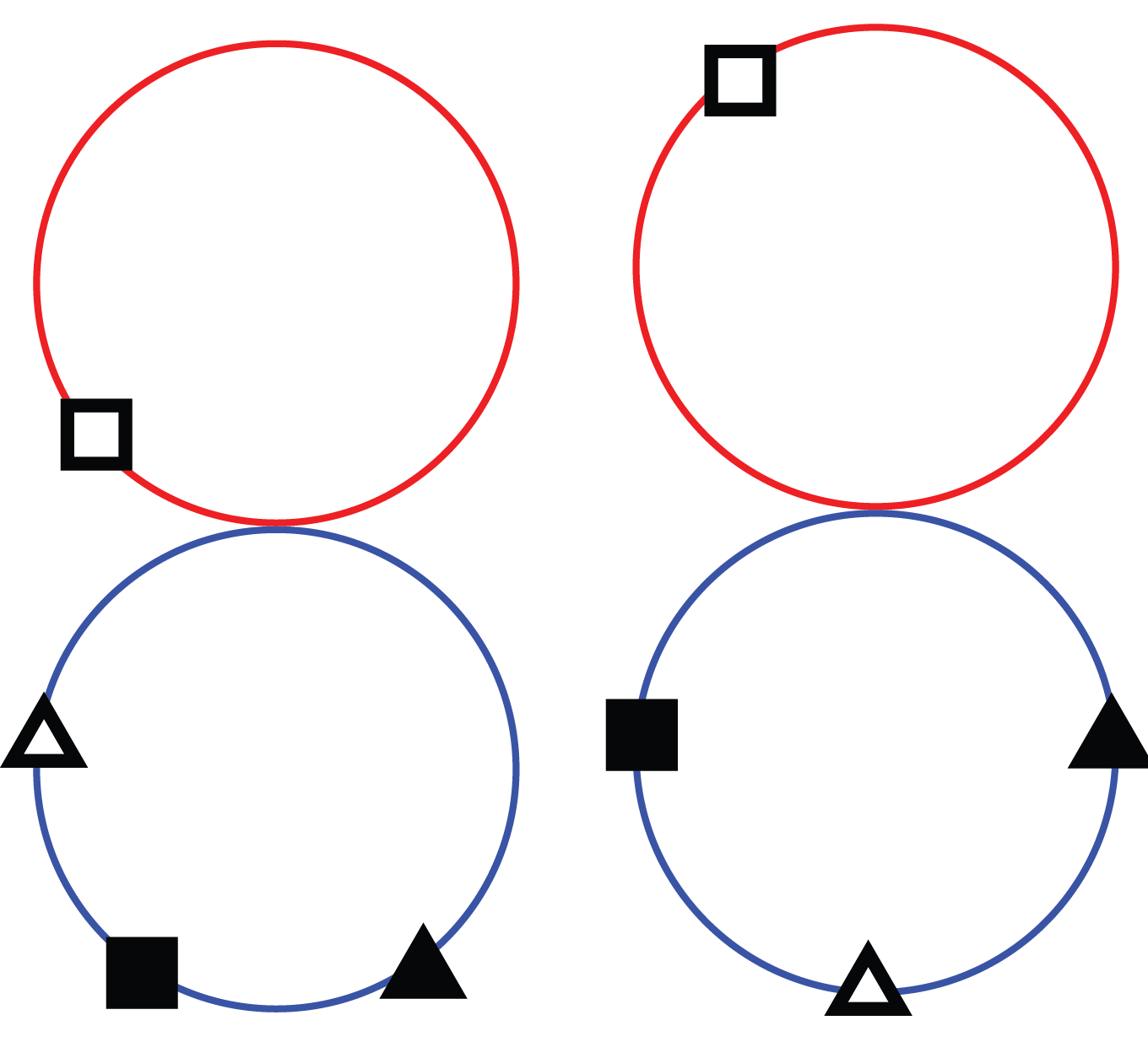}
    \caption{Initial and final positions and their corresponding spine positions.}
    \label{case1}
\end{figure}

Let us consider the initial and final positions shown in figure \ref{case1}.

\begin{figure}[H]
    \centering
    \includegraphics[height=1.2in]{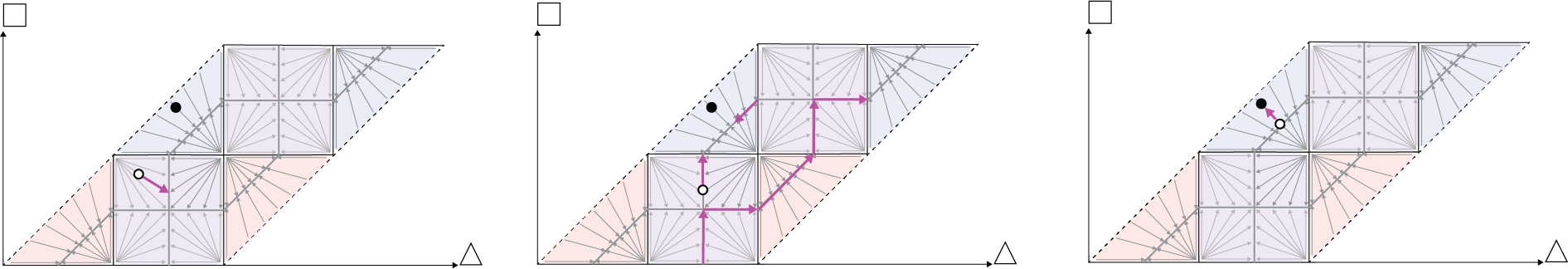}
    \caption{Path in $X$ for case 1.}
    \label{path1}
\end{figure}

The path to move from initial state to final state in the configuration space is shown in figure \ref{path1}.

\begin{figure}[H]
    \centering
    \includegraphics[height=1.2in]{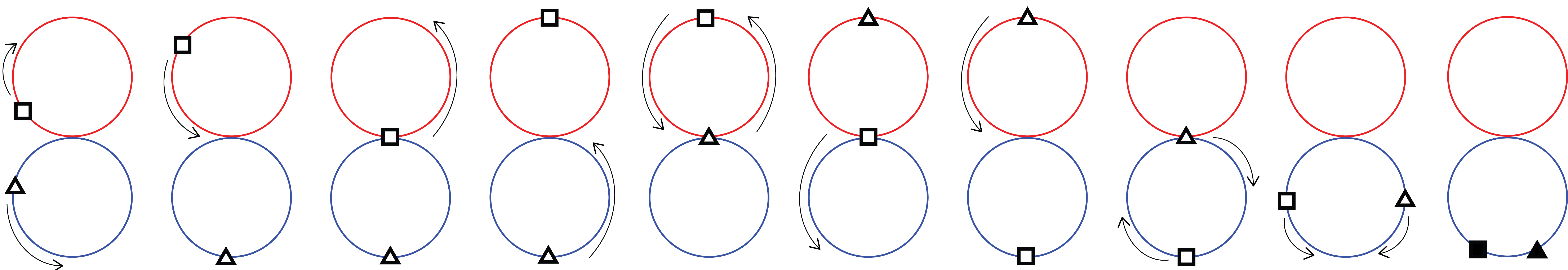}
    \caption{Steps of the algorithm in $\Gamma$  for case 1.}
    \label{gspacealgo}
\end{figure}

The steps of the algorithm to move from initial to final positions in the physical space are shown in figure \ref{gspacealgo}.

\subsubsection{Case scenario 2 }
\begin{figure}[H]
    \centering
    \includegraphics[height=1.5in]{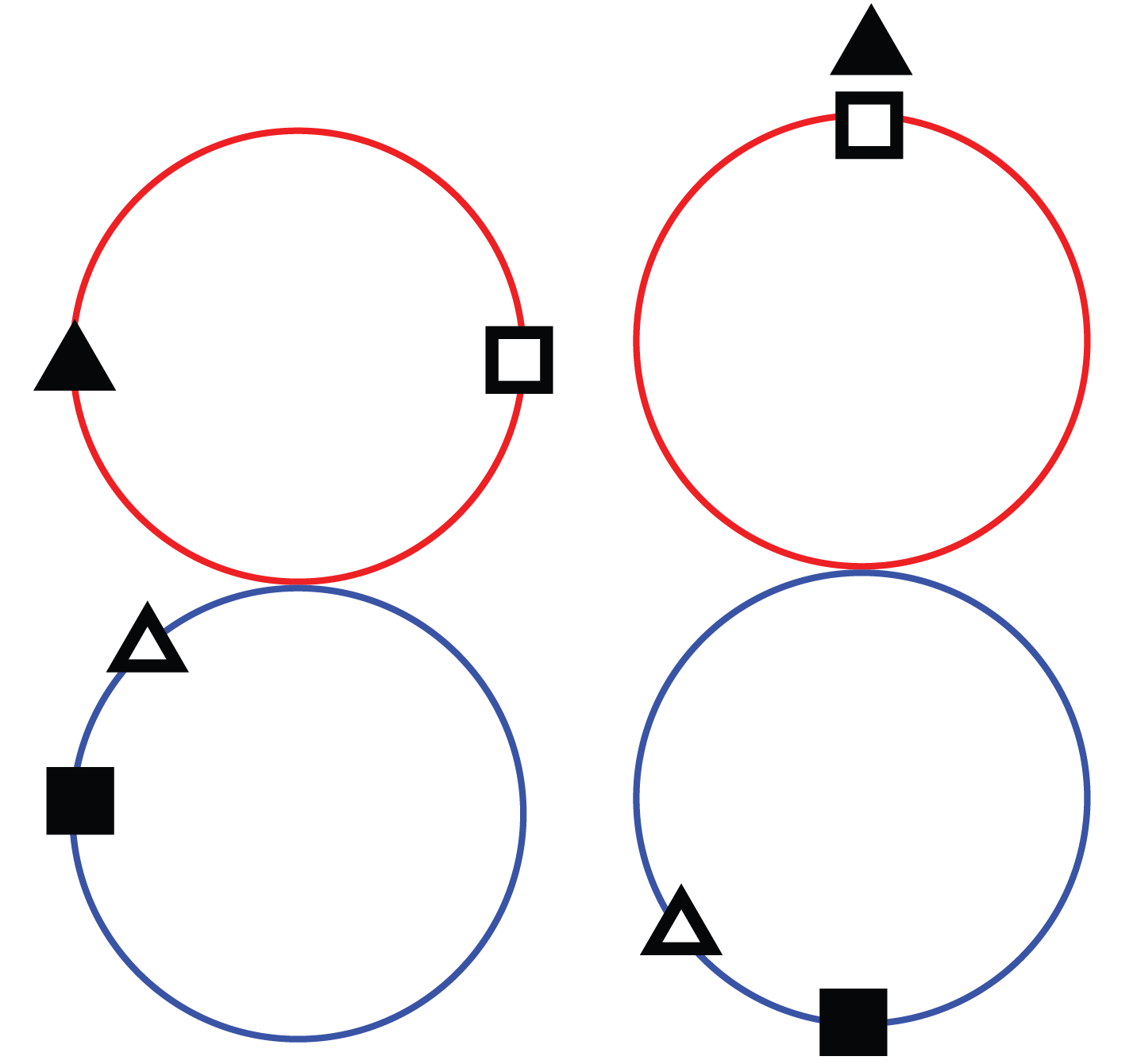}
    \caption{Initial and final positions and their corresponding skeleton positions.}
    \label{case2}
\end{figure}

Let us consider the initial and final positions shown in figure \ref{case2}.

\begin{figure}[H]
    \centering
    \includegraphics[height=1.2in]{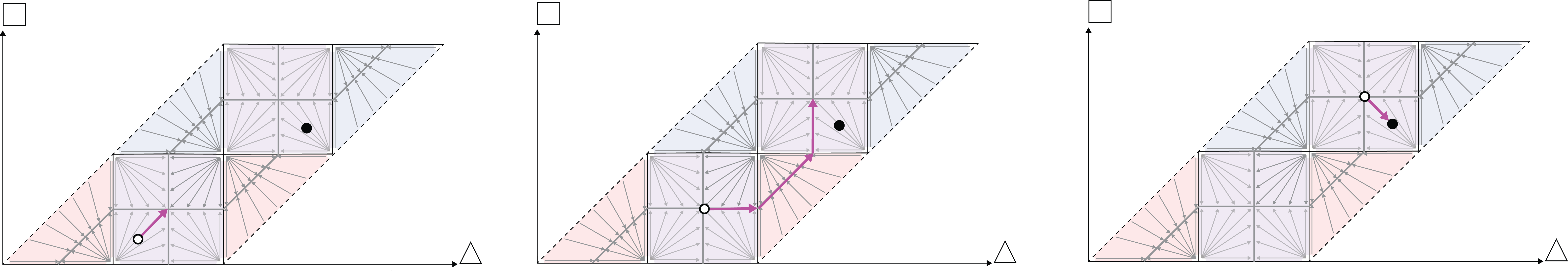}
    \caption{Path in $X$ for case 2.}
    \label{path2}
\end{figure}

The path to move from initial state to final state in the configuration space is shown in figure \ref{path2}.

\begin{figure}[H]
    \centering
    \includegraphics[height=1.3in]{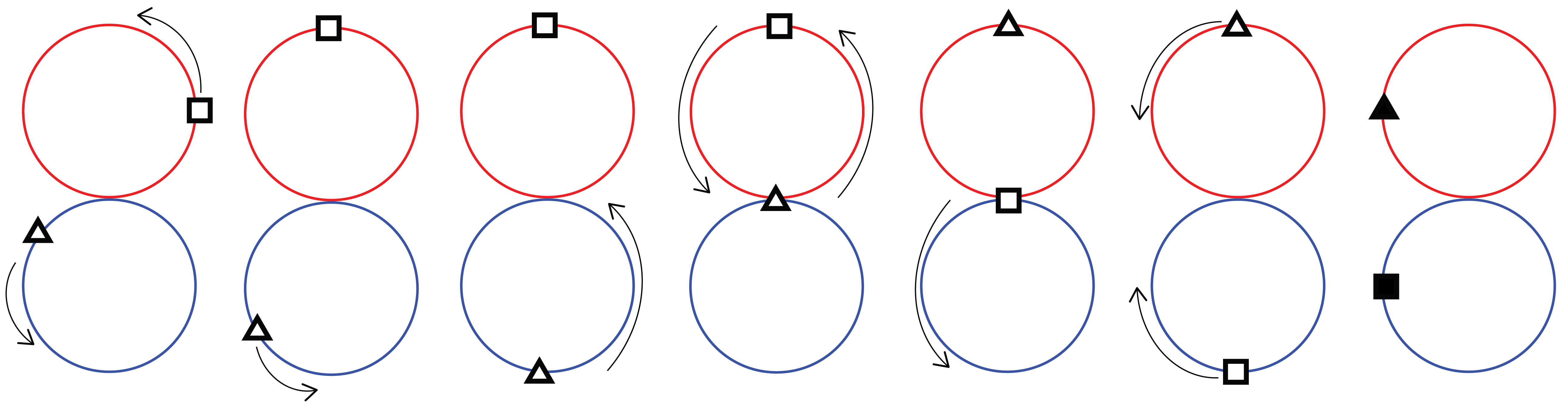}
    \caption{Steps of the algorithm in $\Gamma$  for case 2.}
    \label{s2}
\end{figure}

The steps of the algorithm to move from initial to final positions in the physical space are shown in figure \ref{s2}.

\subsubsection{Case scenario 3 }
\begin{figure}[H]
    \centering
    \includegraphics[height=1.5in]{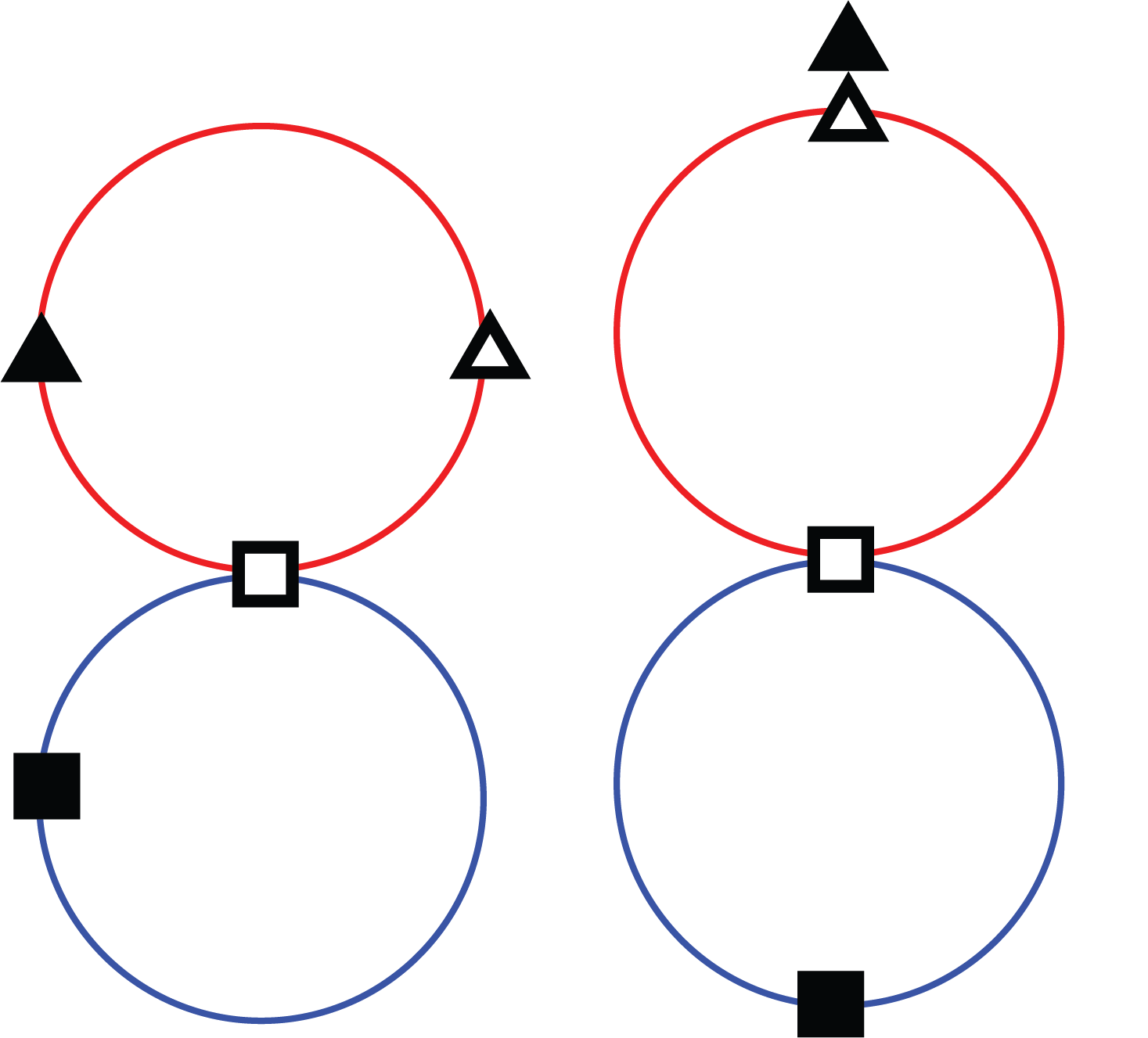}
    \caption{Initial and final positions and their corresponding skeleton positions.}
    \label{case3}
\end{figure}

Let us consider the initial and final positions shown in figure \ref{case3}.

\begin{figure}[H]
    \centering
    \includegraphics[height=1.2in]{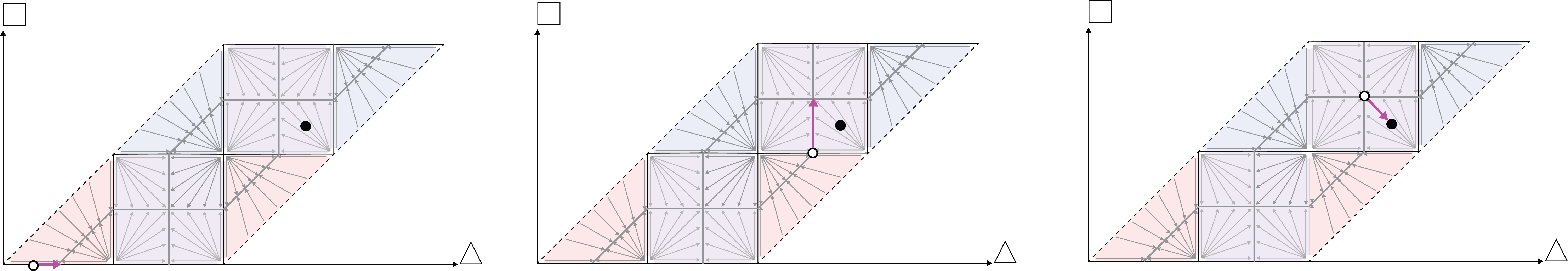}
    \caption{Path in $X$ for case 3.}
    \label{path3}
\end{figure}

The path to move from initial state to final state in the configuration space is shown in figure \ref{path3}.

\begin{figure}[H]
    \centering
    \includegraphics[height=1.5in]{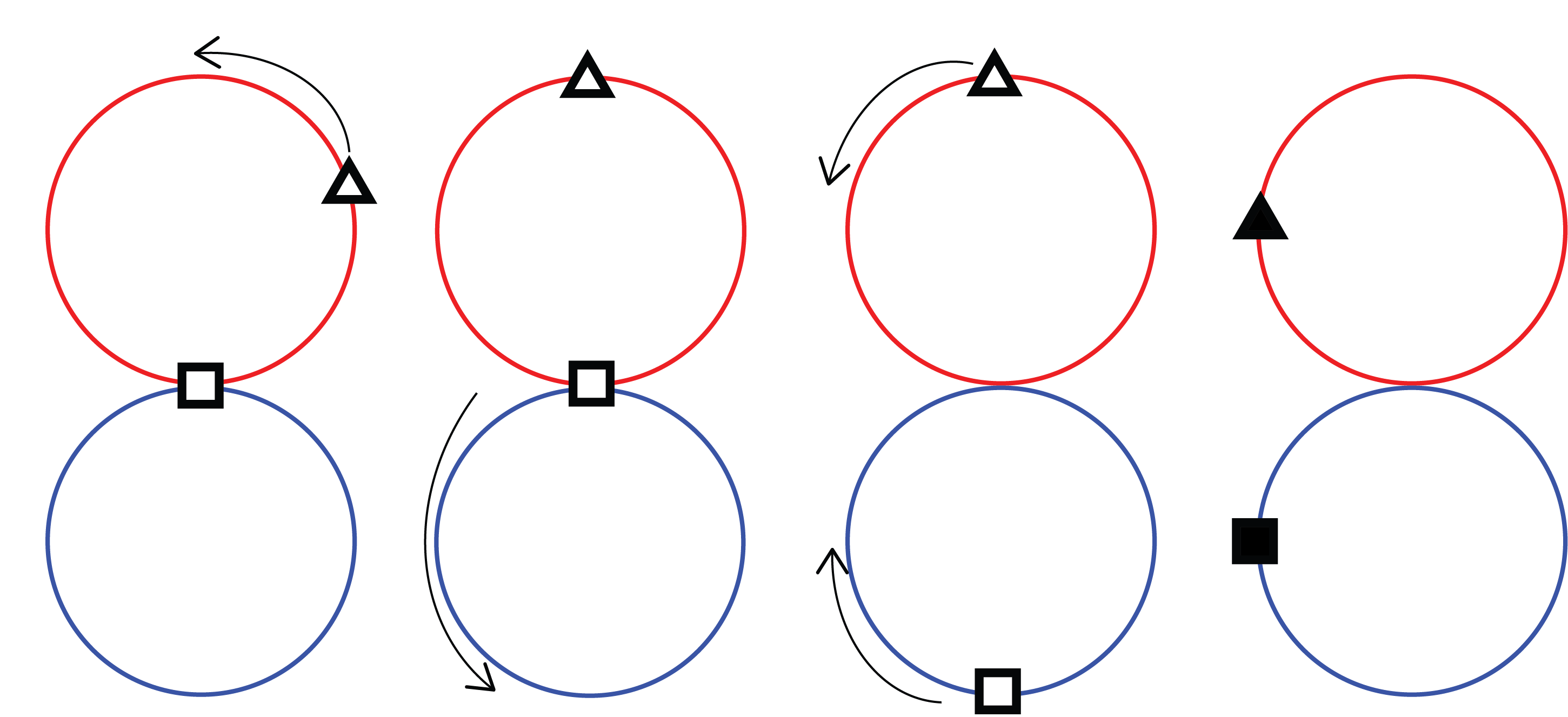}
    \caption{Steps of the algorithm in $\Gamma$  for case 3.}
    \label{s3}
\end{figure}

The steps of the algorithm to move from initial to final positions in the physical space are shown in figure \ref{s3}.

{\footnotesize  
\medskip
\medskip
\vspace*{1mm} 
\noindent{\it Cristian Jardon}\\
Wilbur Wright College\\
4300 N Narragansett Ave\\
Chicago, IL 60634\\
E-mail: {\tt cjardonccc@gmail.com}\\ \\ 

\noindent{\it Brian Sheppard}\\
Harold Washington College\\
30 E. Lake Street\\
Chicago, IL 60601  \\
E-mail: {\tt bshepp@gmail.com}\\ \\ 

\noindent{\it Veet Zaveri}\\
Wilbur Wright College \\
4300 N Narragansett Ave \\
Chicago, IL 60634 \\
E-mail: {\tt veet.zaveri@gmail.com}\\ \\

}

\end{document}